\documentclass[Afour,sagev,times]{sagej}

\usepackage[T1]{fontenc}
\usepackage[utf8]{inputenc}
\usepackage{lmodern}
\usepackage{amsmath,amssymb,amsthm}
\usepackage{booktabs}
\usepackage{multirow}
\usepackage{array}
\usepackage{graphicx}
\usepackage{subcaption}
\usepackage{float}
\usepackage{siunitx}
\usepackage{xcolor}
\usepackage[colorlinks,bookmarksopen,bookmarksnumbered,citecolor=red,urlcolor=red]{hyperref}
\usepackage{url}
\usepackage{enumitem}
\usepackage{longtable}
\usepackage{pdflscape}
\usepackage{setspace}
\usepackage{placeins}
\usepackage{lineno}
\usepackage{algorithm}
\usepackage{algpseudocode}
\usepackage{moreverb}

\newcommand{\paperTitle}{Structural Pass Analysis in Football: Learning Pass Archetypes and Tactical Impact from Spatio-Temporal Tracking Data}
\newcommand{\tiv}{\textit{TIV}}
\newcommand{\lbs}{\textit{LBS}}
\newcommand{\sgm}{\textit{SGM}}
\newcommand{\sdi}{\textit{SDI}}

\def\volumeyear{2026}
\def\journalname{International Journal of Sports Science \& Coaching}
\runninghead{Karaku\c{s}, Arkada\c{s}, and Ey\"{u}po\u{g}lu}

\title{\paperTitle}
\author{Oktay Karaku\c{s}\affilnum{1,2} and Hasan Arkada\c{s}\affilnum{2} and Can Ey\"{u}po\u{g}lu\affilnum{1,3}}
\affiliation{\affilnum{1}Department of Computer Science and Informatics, Cardiff University, Cardiff, United Kingdom\\
\affilnum{2}Dead Ball Analytics Limited, Barry, CF63 2QQ, UK\\
\affilnum{3}Department of Computer Engineering, Turkish Air Force Academy, National Defence University, Istanbul, Türkiye}
\corrauth{Oktay Karaku\c{s}, Department of Computer Science and Informatics, Cardiff University, Cardiff, United Kingdom.}
\email{karakuso@cardiff.ac.uk}

\begin{document}
\begin{abstract}
Passing actions are commonly evaluated through ball progression and attacking outcomes, yet their relationship with the spatial organisation of the opposition provides an additional perspective on passing behaviour. This study introduces an interpretable framework for describing passing structure using synchronised event and tracking data. Three complementary descriptors are proposed: Line Bypass Score (LBS), measuring the difference in goal-side opponent counts between passer and receiver; Space Gain Metric (SGM), comparing their nearest-opponent distances; and Structural Disruption Index (SDI), summarising differences in defensive organisation between pass initiation and a sampled state one second later. Following transformation and standardisation, these descriptors are combined into Tactical Impact Value (TIV), a fixed-weight structural composite. Applied to 41,078 open-play passing events from the 2022 FIFA World Cup, the framework identifies four interpretable profiles: Circulatory, Destabilising, Line-breaking, and Space-expanding. The profiles exhibit distinct structural signatures and different distributions of territorial-crossing and recorded attacking indicators. Team, spatial, player, and passer--receiver analyses further demonstrate how the representation can describe passing behaviour across analytical scales. The framework complements outcome-oriented approaches by connecting individual passing actions with measurable characteristics of opposition positioning and defensive organisation.
\end{abstract}


\maketitle

\section{Introduction}

Passing is fundamental to how football teams organise possession, advance the ball, and create attacking opportunities. Yet passes with similar starting and ending locations can differ considerably in their relationship with the opposition. One may bypass several defenders, another may reach a player with greater space to receive, while a third may occur alongside substantial differences in defensive organisation. Understanding these distinctions requires examining not only where the ball travels but also the spatial circumstances surrounding the action.

The increasing availability of event and tracking data has expanded the possibilities for studying these characteristics. Event-based approaches describe on-ball actions and their attacking consequences. Expected goals (xG) models evaluate shooting opportunities \cite{lucey2014quality}, while Valuing Actions by Estimating Probabilities (VAEP) and expected threat (xT) provide established perspectives on action value and attacking progression \cite{decroos2019actions,singh2019xt}. Tracking data complement these approaches by capturing player positioning and movement, supporting analyses of spatial control, passing options, and the creation and exploitation of space \cite{spearman2018beyond,fernandez2018wide,rahimian2025pass}. Together, these developments have substantially improved our ability to examine how attacking actions unfold within the organisation of both teams.

An important analytical opportunity lies in describing the structural characteristics of individual passes independently of their eventual outcomes. Two passes reaching comparable areas may bypass different numbers of opponents or offer markedly different receiving conditions. Similarly, differences in defensive organisation may accompany passes that appear relatively ordinary in event records. Representing these characteristics jointly could provide a more informative account of passing behaviour and help distinguish spatial patterns that conventional event descriptions do not fully capture. Previous work by Karakuş and Arkadaş \cite{karakus2026lbp} examined line-breaking passes using synchronised event and tracking data, employing spatial segmentation to investigate their relationship with defensive organisation. The present study extends this perspective by considering several complementary structural properties within a common framework.

We introduce three descriptors capturing distinct aspects of the relationship between passing and opposition positioning. The Line Bypass Score (LBS) characterises positional bypass through the signed difference in goal-side opponent counts between passer and receiver, while the Space Gain Metric (SGM) compares their nearest-opponent distances using a floored passer reference. The Structural Disruption Index (SDI) adds a temporal perspective by summarising absolute differences in defensive-shape characteristics between pass initiation and a sampled state approximately one second later. Together, these measurements provide a multidimensional representation of passing structure. We further combine their transformed and standardised values into Tactical Impact Value (TIV), a fixed-weight composite designed to summarise the measured structural characteristics rather than predict attacking outcomes.

The framework is applied to 41,078 open-play passing events across all 64 matches of the 2022 FIFA World Cup. Unsupervised clustering identifies four operational passing profiles---Circulatory, Destabilising, Line-breaking, and Space-expanding---which are examined through their structural signatures and relationships with territorial-crossing and recorded attacking indicators. We subsequently explore how the same representation supports comparisons between teams, across pitch locations, among individual players, and within recurring passer--receiver combinations.

The contribution of this study is threefold: it establishes complementary tracking-derived descriptors of passing structure, organises their combinations into an interpretable taxonomy and composite representation, and demonstrates their application across several levels of football analysis. By connecting the spatial characteristics of individual passes with broader patterns of observed behaviour, the framework offers a complementary perspective on how passing interacts with opposition organisation.

\section{Related Work}
\label{sec:related_work}

\subsection{Action valuation and pass evaluation}

Football action valuation has developed several approaches for quantifying the contribution of individual events to attacking performance. Expected goals (xG) models assess shooting opportunities \cite{lucey2014quality}, while Valuing Actions by Estimating Probabilities (VAEP) evaluates changes in scoring and conceding probabilities \cite{decroos2019actions}. Expected threat (xT) provides a complementary location-based representation of attacking progression \cite{singh2019xt}. Related research examines the contribution of passes to chance creation \cite{bransen2019passes} and incorporates tracking information into expected possession value (EPV) models \cite{fernandez2021framework}. Contextual and counterfactual approaches further extend attacking evaluation by accounting for player characteristics and alternative situations \cite{hewitt2023player,scholtes2024bayes,mahmudlu2025counterfactual}.

A related research direction concerns the difficulty and feasibility of passing. By incorporating contextual and spatial information, passing models examine the circumstances associated with successful execution rather than focusing exclusively on subsequent attacking consequences \cite{szczepanski2016beyond,anzer2022expected}. Together, these approaches provide established perspectives on action value, progression, and execution. The present study complements them by examining the structural relationships between passers, receivers, and opposition players, independently of an estimated attacking outcome.

\subsection{Tracking-data-based spatial analysis}

Player-tracking data have enabled football to be studied as a continuously evolving spatial system, supporting investigations of movement, interaction, and collective organisation \cite{gudmundsson2017spatio}. Research on passing probability and spatial control has demonstrated how player positioning influences available passing opportunities \cite{spearman2017physics,spearman2018beyond}. Related work examines the creation and exploitation of space through interactions between attacking and defending players \cite{fernandez2018wide,martens2021space}.

Tracking information also supports the analysis of collective defensive organisation, including formations and positional relationships \cite{sotudeh2025formation}. These developments establish the relevance of spatial context when examining passing behaviour. Building on this perspective, our framework describes positional bypass, relative receiving space, and sampled differences in defensive organisation through complementary quantitative descriptors.

\subsection{Passing against defensive structure}

Passing can be understood as an interaction with opposition positioning rather than solely as ball movement between two locations. Previous research has examined pass effectiveness in relation to defensive organisation \cite{goes2019not} and classified passing actions using spatio-temporal and computational-geometric information \cite{horton2014classification}. More specialised studies investigate penetrative and line-breaking passes, highlighting the importance of defensive positioning when characterising progression \cite{rahimian2022penetrative,karakus2026lbp}.

Passing decisions have also been studied through models of receiver selection, available options, feasibility, and reception outcomes \cite{eigenrauch2024predicting,rahimian2023passreceiver,rahimian2025pass,anzer2022expected}. These approaches demonstrate how the configuration of players influences passing opportunities and execution. Our framework addresses a complementary question by jointly describing the positional relationships and defensive-shape characteristics associated with observed passing actions.

\subsection{Unsupervised learning and tactical pattern discovery}

Unsupervised learning provides a means of identifying recurring patterns in football data by organising observations according to similarities in their measured characteristics. Previous research has demonstrated its application to tactical pattern discovery and interpretation \cite{decroos2018automatic}.

This perspective is particularly relevant to structural passing analysis because individual actions may exhibit different combinations of positional bypass, receiving space, and defensive-shape variation. Rather than treating these characteristics independently, clustering allows their recurring combinations to be examined through interpretable profiles. The present study applies this approach to construct an operational four-profile taxonomy, connecting pass-level structural measurements with comparisons across teams, players, and pitch locations.

\section{Data}
\label{sec:data}

This study uses synchronised event and tracking data from the 2022 FIFA World Cup, publicly released by PFF FC\footnote{\url{https://www.blog.fc.pff.com/blog/pff-fc-release-2022-world-cup-data}}. The dataset covers all 64 tournament matches and combines timestamped on-ball events with the spatial positions of players and the ball, allowing passing actions to be examined in relation to opposition organisation.

\subsection{Tracking and Event Data}

The dataset comprises structured JSON files containing event records, tracking observations, match metadata, and player information. Event records identify actions and their timestamps, coordinates, and contextual attributes, while tracking observations provide two-dimensional player and ball positions at 29.97 Hz. Match metadata specify pitch dimensions, coordinate orientation, and frame rate; roster information links player identifiers to recorded participants and roles.

Events are aligned with tracking observations using the match period and timestamp. For each candidate pass, the nearest available frame within the corresponding period is selected, with ties resolved in favour of the earlier frame. Events outside the available tracking range are associated with the nearest boundary frame. Player positions are retrieved from the smoothed tracking collection when available, otherwise from the corresponding unsuffixed collection. These aligned configurations provide the spatial observations required for the proposed descriptors.

\subsection{Pass Extraction}

The analysis retains open-play passing events with identifiable passers and intended receivers, excluding set-piece deliveries such as corners, throw-ins, and free kicks. Each retained event is associated with its selected tracking configuration, from which the positions of the passer, receiver, and opposition players are obtained. The resulting sample supports consistent calculation of the structural passing descriptors across the tournament.

\subsection{Dataset Statistics}

The analytical sample contains 41,078 open-play passing events from 64 matches involving 32 national teams, averaging 641.84 analysed passes per match. Tracking observations contain the positions of 22 players per frame at 29.97 Hz. Table~\ref{tab:dataset_summary} summarises the principal dataset characteristics.

\begin{table}[ht]
    \centering
    \caption{Summary of the dataset used for structural pass analysis.}
    \label{tab:dataset_summary}
    \resizebox{\columnwidth}{!}{%
    \begin{tabular}{lc}
        \toprule
        Quantity & Value \\
        \midrule
        Matches & 64 \\
        Teams & 32 \\
        Total analysed passes & 41,078 \\
        Mean passes per match & 641.84 \\
        Tracking frame rate & 29.97 \\
        Players tracked per frame & 22 \\
        Competition & FIFA World Cup 2022 \\
        \bottomrule
    \end{tabular}%
    }
\end{table}

\section{Methodology}
\label{sec:method}

\begin{figure*}[t]
    \centering
    \includegraphics[width=\linewidth]{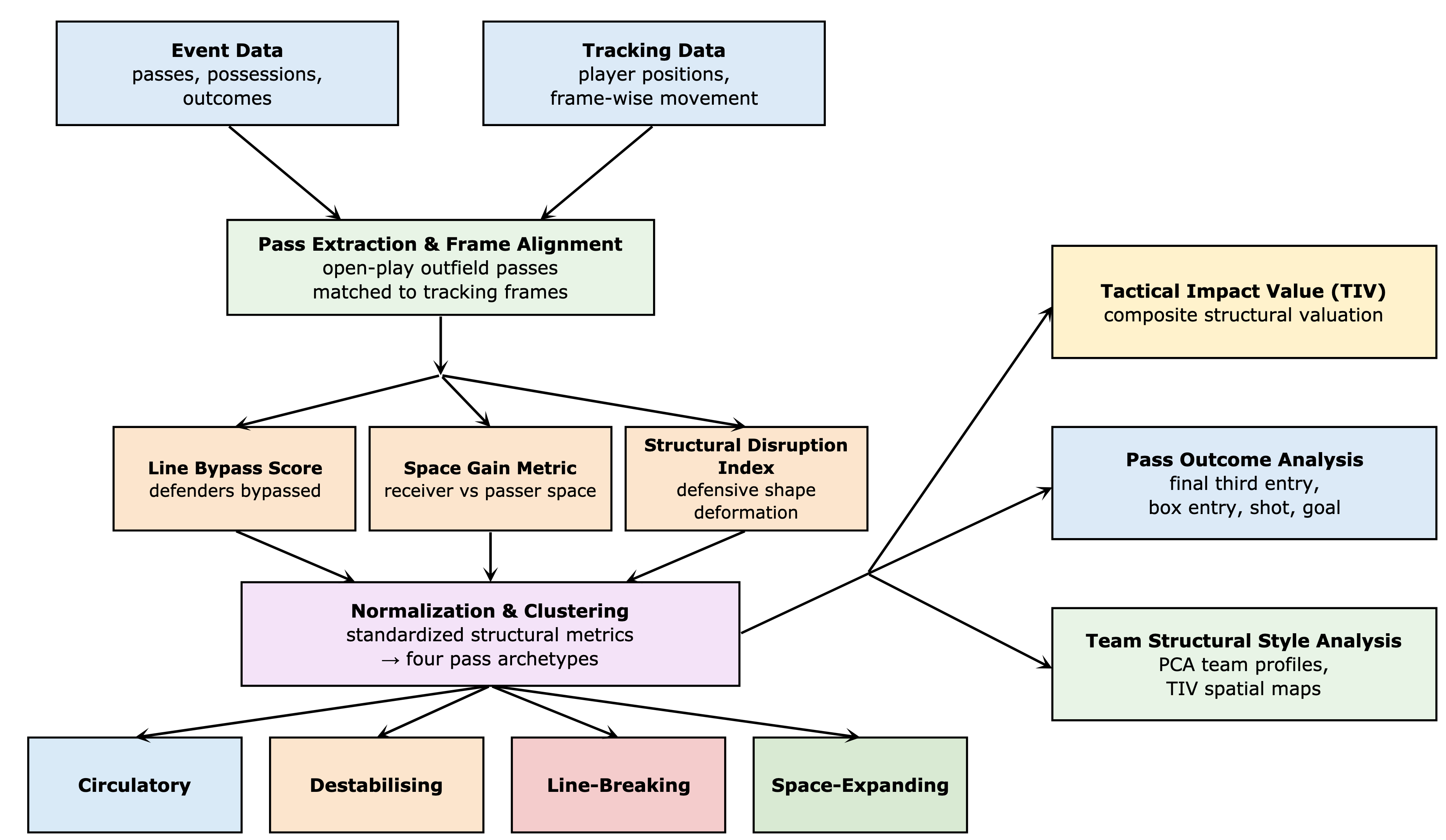}
    \caption{Analytical workflow integrating event and tracking data, structural descriptors, TIV, pass clustering, and subsequent team- and player-level analyses.}
    \label{fig:method_pipeline}
\end{figure*}

The proposed framework describes passing structure through three complementary characteristics: positional bypass, relative receiving space, and differences in defensive organisation. These are quantified using the Line Bypass Score (LBS), Space Gain Metric (SGM), and Structural Disruption Index (SDI). Following transformation and standardisation, the descriptors support two analytical representations: Tactical Impact Value (TIV), a weighted structural composite, and an unsupervised taxonomy of recurring passing profiles. Figure~\ref{fig:method_pipeline} summarises the analytical workflow.

\subsection{Conceptual Structural Perspective}

\begin{figure}[t]
    \centering
    \includegraphics[width=\columnwidth]{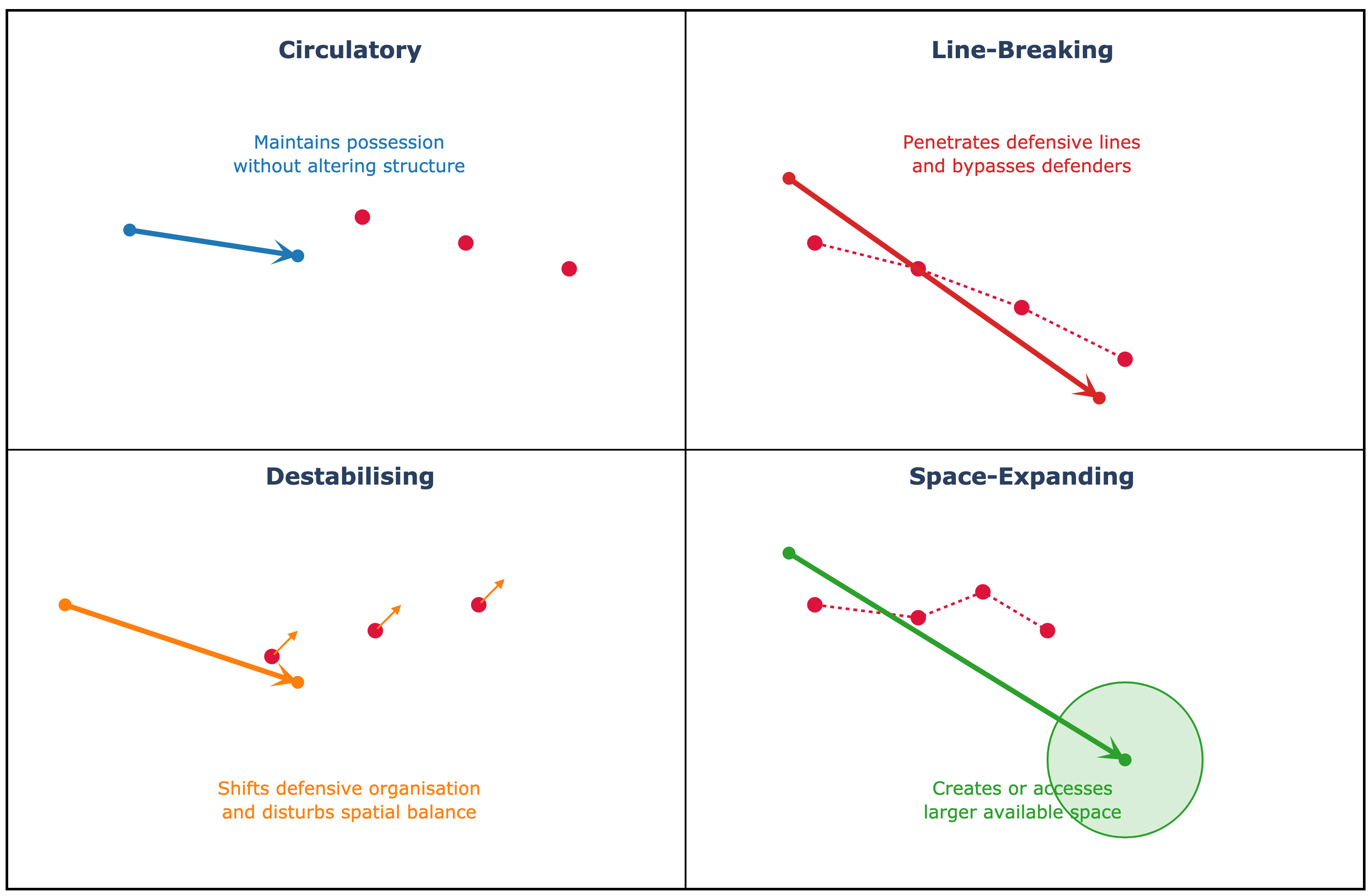}
    \caption{Conceptual illustration of the four structural passing profiles: Circulatory, Destabilising, Line-breaking, and Space-expanding.}
    \label{fig:conceptual_perspective}
\end{figure}

Passing actions can differ in their relationship with opposition positioning even when their recorded event characteristics are similar. We consider four interpretable structural profiles: Circulatory, Destabilising, Line-breaking, and Space-expanding (Figure~\ref{fig:conceptual_perspective}). These profiles provide a conceptual interpretation of passing structure, subsequently examined through clustering the proposed descriptors.

\subsection{Structural Representation of a Pass}

Consider an eligible passing event $p_i$ at time $t_i$. The event is associated with the nearest tracking frame within its match period, resolving equal temporal distances in favour of the earlier frame and assigning out-of-range events to the nearest boundary frame. Player coordinates are obtained from the nonempty smoothed tracking collection where available, otherwise from the corresponding unsuffixed collection.

Let $\mathbf{x}_s=(x_s,y_s)$ and $\mathbf{x}_r=(x_r,y_r)$ denote the tracked positions of the passer and receiver. The opposition configuration at pass initiation is

\[
O_i^{(0)}
=
\{\mathbf{d}^{(0)}_1,\ldots,\mathbf{d}^{(0)}_{n_0}\}.
\]

A second configuration, $O_i^{(1)}$, is retrieved from the nearest tracking frame to $t_i+1.0$ seconds within the same match period. Opposition players are identified using team membership and the available positional visibility and confidence criteria. LBS and SGM use the pass-time configuration, whereas SDI compares the two sampled defensive states.

\subsection{Structural Pass Metrics}

\begin{figure*}[t]
    \centering
    \includegraphics[width=\linewidth]{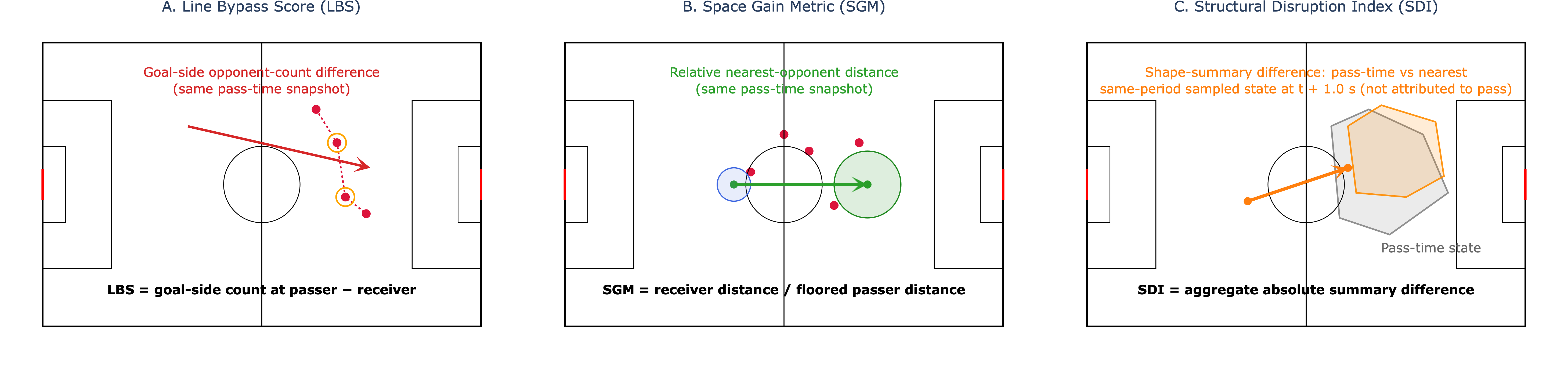}
    \caption{Illustration of the structural descriptors. LBS compares goal-side opponent counts at passer and receiver coordinates within the same pass-time configuration; SGM compares nearest-opponent distances. SDI measures differences between the pass-time defensive configuration and the nearest sampled state approximately one second later, rather than an identified reception state.}
    \label{fig:metric_illustration}
\end{figure*}

Figure~\ref{fig:metric_illustration} illustrates the spatial references used by the three descriptors. Their mathematical definitions and interpretations are developed below.

\subsubsection{Line Bypass Score (\lbs)}

LBS measures positional bypass through the signed difference between the numbers of opponents located goal-side of the passer and receiver. Let $a_i=1$ indicate an attacking direction towards increasing $x$ coordinates and $a_i=-1$ the opposite direction. For opponent $j$, let $x_j^{(0)}$ denote the horizontal coordinate in the pass-time configuration.

The corresponding goal-side opponent counts are

\begin{align}
B_s &=
\begin{cases}
\displaystyle\sum_{j=1}^{n_0}\mathbf{1}[x_j^{(0)}>x_s], & a_i=1,\\[2pt]
\displaystyle\sum_{j=1}^{n_0}\mathbf{1}[x_j^{(0)}<x_s], & a_i=-1,
\end{cases} \\
B_r &=
\begin{cases}
\displaystyle\sum_{j=1}^{n_0}\mathbf{1}[x_j^{(0)}>x_r], & a_i=1,\\[2pt]
\displaystyle\sum_{j=1}^{n_0}\mathbf{1}[x_j^{(0)}<x_r], & a_i=-1.
\end{cases}
\end{align}

The descriptor is defined as

\[
\lbs(p_i)=B_s-B_r.
\]

Positive values indicate fewer opponents positioned goal-side of the receiver than the passer, with larger values corresponding to greater positional bypass under this criterion. Zero denotes equal counts, while negative values indicate more opponents goal-side of the receiver. Both counts use the same defensive snapshot; LBS therefore describes positional relationships rather than defender movement or reconstructed formation lines.

\subsubsection{Space Gain Metric (\sgm)}

SGM describes relative receiving space through the nearest-opponent distances of the passer and receiver. Using the pass-time opposition configuration, these distances are

\[
d_s=\min_{\mathbf{d}\in O_i^{(0)}}\|\mathbf{x}_s-\mathbf{d}\|_2,
\qquad
d_r=\min_{\mathbf{d}\in O_i^{(0)}}\|\mathbf{x}_r-\mathbf{d}\|_2.
\]

The passer's distance is subject to a minimum reference value of 1.5 distance units:

\[
\widehat{d}_s=\max(1.5,d_s),
\qquad
\sgm(p_i)=\frac{d_r}{\widehat{d}_s}.
\]

Values above one indicate that the receiver's nearest-opponent distance exceeds the adjusted passer reference; values below one indicate the opposite relationship. SGM thus captures relative defensive proximity at pass initiation rather than estimating the probability of successful reception or comprehensive spatial control.

\subsubsection{Structural Disruption Index (\sdi)}

SDI summarises differences in the defending team's spatial organisation between the pass-time configuration and the nearest sampled state approximately one second later. Five geometric characteristics are considered: width, depth, horizontal and vertical centroid coordinates, and convex-hull area.

For each state $q\in\{0,1\}$, let defender $j$ have coordinates $(x_j^{(q)},y_j^{(q)})$ within $O_i^{(q)}$. The defensive-shape characteristics are defined as

\[
W^{(q)}=\max_j y_j^{(q)}-\min_j y_j^{(q)},
\]

\[
D^{(q)}=\max_j x_j^{(q)}-\min_j x_j^{(q)},
\]

\[
C_x^{(q)}=\frac{1}{n_q}\sum_{j=1}^{n_q}x_j^{(q)},
\]

\[
C_y^{(q)}=\frac{1}{n_q}\sum_{j=1}^{n_q}y_j^{(q)},
\]

\[
A^{(q)}=\operatorname{Area}\!\left(\operatorname{ConvexHull}(O_i^{(q)})\right).
\]

Their absolute differences are

\[
\begin{aligned}
\Delta_W &= |W^{(1)}-W^{(0)}|, &
\Delta_D &= |D^{(1)}-D^{(0)}|, \\
\Delta_{C_x} &= |C_x^{(1)}-C_x^{(0)}|, &
\Delta_{C_y} &= |C_y^{(1)}-C_y^{(0)}|, \\
\Delta_A &= |A^{(1)}-A^{(0)}|.
\end{aligned}
\]

The Structural Disruption Index is their unnormalised sum:

\[
\sdi(p_i)=\Delta_W+\Delta_D+\Delta_{C_x}+\Delta_{C_y}+\Delta_A.
\]

Higher values indicate larger aggregate differences across the sampled defensive-shape characteristics. The five components enter with unit weights and retain their original numerical scales. As absolute differences are used, SDI captures the magnitude rather than the direction of variation. The sampled comparison does not establish that the passing action caused the observed differences.

Table~\ref{tab:metric_definitions} summarises the descriptors and their football interpretations.

\begin{table*}[t]
\centering
\caption{Definitions and interpretations of the three structural passing descriptors.}
\label{tab:metric_definitions}
\small
\begin{tabular}{p{3.5cm} p{6.8cm} p{5.2cm}}
\toprule
Metric & Definition & Tactical interpretation \\
\midrule
Line Bypass Score (\lbs) &
Signed difference between pass-time goal-side opponent counts at the tracked passer and receiver x-coordinates. &
Positive values indicate fewer opponents goal-side at the receiver coordinate; negative values are possible. \\
Space Gain Metric (\sgm) &
Ratio of receiver to floored passer nearest-opponent distance, using the same pass-time opponent snapshot. &
Values above 1 indicate a larger receiver nearest-opponent distance relative to the floored passer distance. \\
Structural Disruption Index (\sdi) &
Unnormalised sum of absolute changes in defensive width, depth, centroid x/y, and convex-hull area between pass-time and one-second-later sampled states. &
Higher values indicate larger aggregate differences across the five shape summaries; the total does not indicate change direction. \\
\bottomrule
\end{tabular}
\end{table*}

\subsection{Metric Transformations and Tactical Impact Value}

Because the descriptors have different numerical scales, they are transformed before standardisation. LBS is divided by its maximum absolute value across the analysed sample, while SGM and SDI undergo logarithmic transformations:

\begin{align}
L_i^{*} &= \frac{\lbs(p_i)}{\max_k|\lbs(p_k)|}, \\
G_i^{*} &= \log(1+\sgm(p_i)), \\
S_i^{*} &= \log(1+\sdi(p_i)).
\end{align}

A \texttt{StandardScaler} is fitted to the analysed dataset and applied separately to the transformed descriptors, producing standardised features $Z_{L,i}$, $Z_{G,i}$, and $Z_{S,i}$. These features are used for both TIV and clustering.

TIV is defined as the fixed-weight composite

\[
\tiv(p_i)=0.35Z_{L,i}+0.25Z_{G,i}+0.40Z_{S,i}.
\]

The weights assign relative contributions of 0.35 to positional bypass, 0.25 to relative space, and 0.40 to defensive-shape variation. No additional rescaling is applied to the weighted sum. TIV therefore summarises the combined structural measurements rather than representing an estimated scoring probability or expected possession value.

\subsection{Pass Clustering and Structural Typology}

To identify recurring combinations of structural characteristics, each pass is represented by the standardised feature vector

\[
\mathbf{z}_i =
\left(
Z_{L,i},\,
Z_{G,i},\,
Z_{S,i}
\right)^\top.
\]

K-means clustering partitions these observations by minimising the within-cluster sum of squared Euclidean distances:

\[
\arg\min_{C}
\sum_{k=1}^{K}
\sum_{\mathbf{z}_i \in C_k}
\|\mathbf{z}_i - \boldsymbol{\mu}_k\|^2,
\]

where $C_k$ denotes cluster $k$ and $\boldsymbol{\mu}_k$ its centroid. The implementation uses $K=4$, \texttt{random\_state=42}, and \texttt{n\_init=50}.

The selected partition is interpreted through four operational profiles: Circulatory, Destabilising, Line-breaking, and Space-expanding. Circulatory passes exhibit comparatively modest structural measurements; Destabilising passes have larger SDI values; Line-breaking passes show stronger positive positional bypass; and Space-expanding passes are distinguished by greater relative receiving space.

In the current implementation, cluster identifiers are mapped as follows: cluster 0 to Circulatory, cluster 1 to Line-breaking, cluster 2 to Destabilising, and cluster 3 to Space-expanding. These identifiers are operational labels rather than ordinal rankings.

Cluster characteristics are examined using raw metric summaries and visualisations. Figure~\ref{fig:structural_map} presents the observations in the original three-dimensional metric space, while Figure~\ref{fig:structural_projection} provides a two-component principal component analysis (PCA) projection of the raw descriptors.

\subsection{Territorial-entry indicators and outcome flags}

The analysis incorporates binary indicators of final-third and penalty-area coordinate crossings, together with stored shot and goal flags. The final-third crossing criterion is

\[
x_{i,s}\leq17.5
\quad\text{and}\quad
x_{i,e}>17.5,
\]

while box entry requires

\[
x_{i,s}\leq36
\quad\text{and}\quad
x_{i,e}>36
\quad\text{and}\quad
|y_{i,e}|<20.
\]

Here, $(x_{i,s},y_{i,s})$ and $(x_{i,e},y_{i,e})$ denote the supplied starting and ending coordinates of pass $i$. These indicators are calculated directly from the specified coordinate-crossing conditions.

Within each archetype, the corresponding rates are obtained by averaging the binary territorial indicators. The stored shot and goal flags are summarised in the same manner, providing descriptive comparisons between structural passing profiles and the additional recorded variables.

\subsection{Team, spatial, player, and pair summaries}

The structural representation is subsequently aggregated across teams, pitch locations, players, and passer--receiver combinations.

Team profiles are constructed from the proportions of analysed passes assigned to each archetype. Two-component PCA supports visual comparison of the resulting team-level profiles, alongside additional summaries including cumulative TIV.

Spatial distributions are examined by aggregating pass origins and destinations into a $12\times8$ coordinate grid and calculating mean TIV within each bin. An $8\times4$ destination grid is additionally used for comparisons involving eight selected national teams.

Player-level analysis considers individuals averaging at least 40 analysed passes per match. The 15 eligible players with the highest cumulative TIV are displayed alongside their mean analysed passes per match, mean TIV, component metrics, and archetype proportions. Cumulative TIV is the sum of pass-level values, while mean TIV is their arithmetic average.

Finally, passer--receiver analysis focuses on the ten players with the highest cumulative TIV and retains combinations involving at least ten analysed passes. For each eligible pair, the difference between its mean TIV and the passer's overall mean is calculated:

\[
\Delta TIV_{p\rightarrow r}
=
\overline{TIV}_{p\rightarrow r}
-
\overline{TIV}_{p}.
\]

The ten combinations with the largest positive differences are displayed together with their passing statistics and recorded attacking indicators. These complementary summaries allow the same structural representation to be examined across different levels of football analysis.

\section{Results}
\label{sec:results}

The analysis examines the structural characteristics of 41,078 passing events from the 2022 FIFA World Cup, beginning with the four passing archetypes and their relationships with territorial and attacking indicators. Subsequent analyses explore how these characteristics vary across teams, pitch locations, players, and recurring passing relationships.

\subsection[Representative examples of structural pass archetypes]{Representative examples of structural pass\\archetypes}

\begin{figure}[t]
\centering

\begin{subfigure}{0.41\columnwidth}
    \centering
    \includegraphics[width=\linewidth]{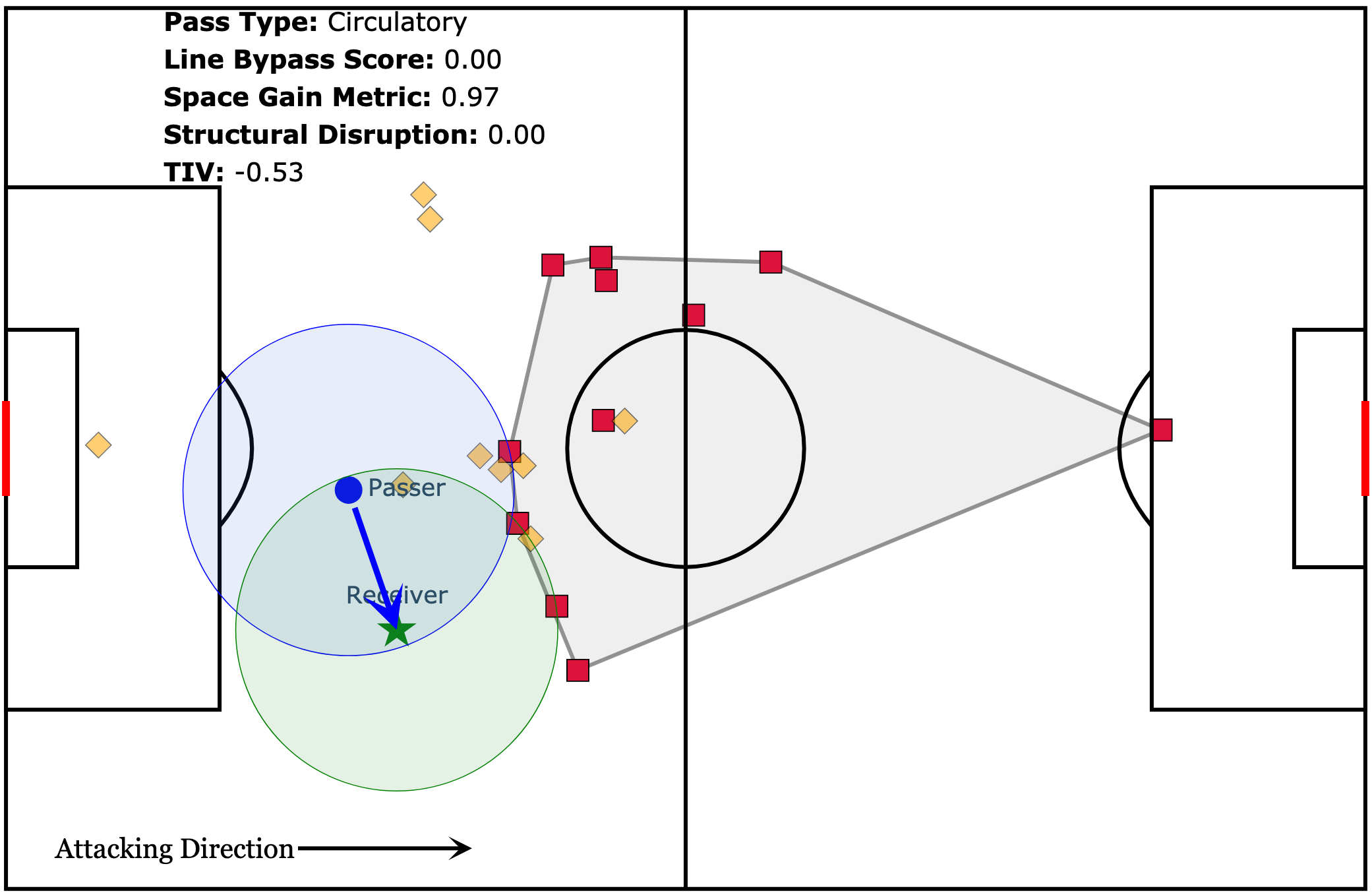}
    \caption{Circulatory pass}
\end{subfigure}
\hfill
\begin{subfigure}{0.55\columnwidth}
    \centering
    \includegraphics[width=\linewidth]{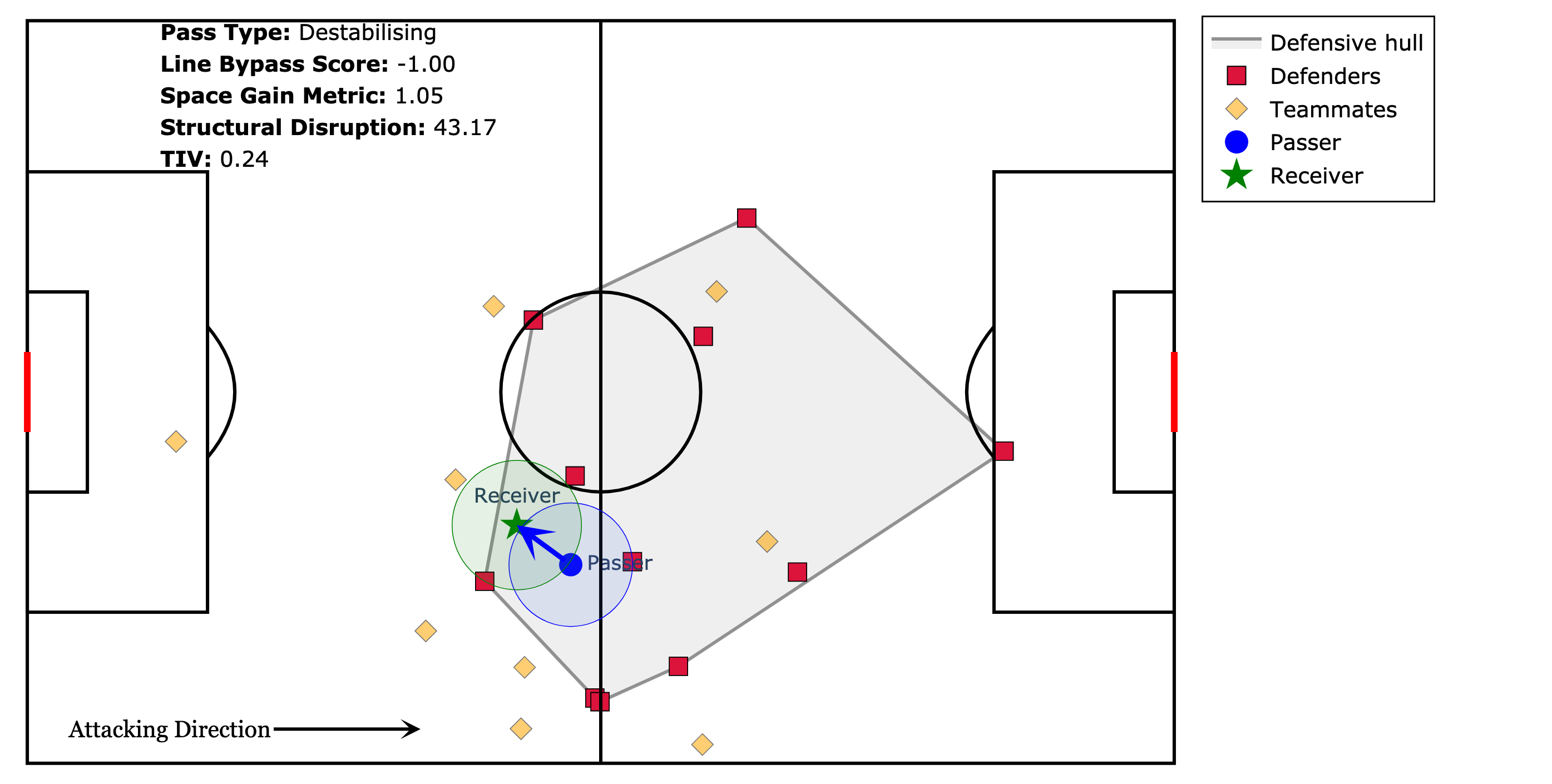}
    \caption{Destabilising pass}
\end{subfigure}

\vspace{0.3cm}

\begin{subfigure}{0.41\columnwidth}
    \centering
    \includegraphics[width=\linewidth]{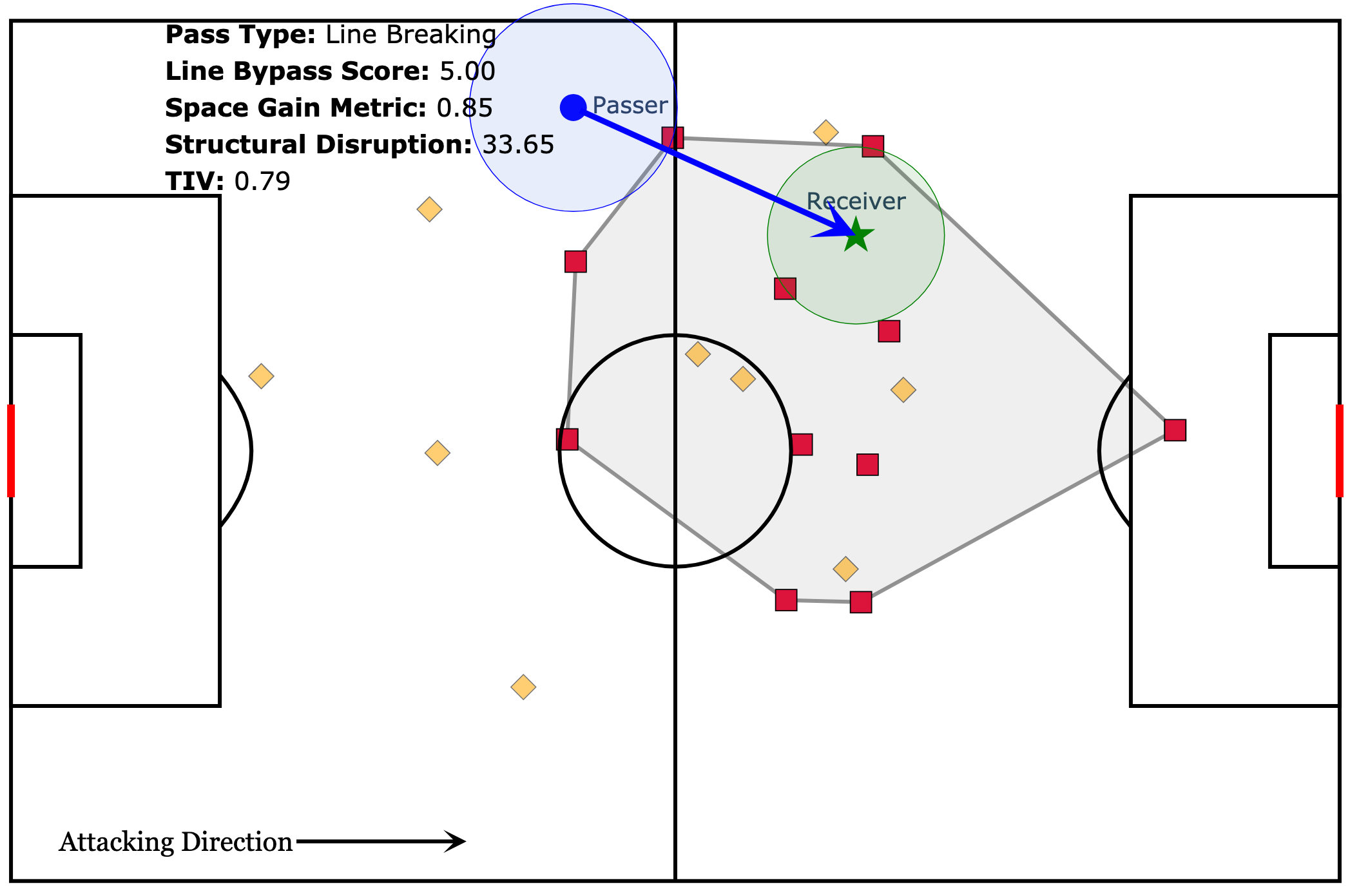}
    \caption{Line-breaking pass}
\end{subfigure}
\hfill
\begin{subfigure}{0.55\columnwidth}
    \centering
    \includegraphics[width=\linewidth]{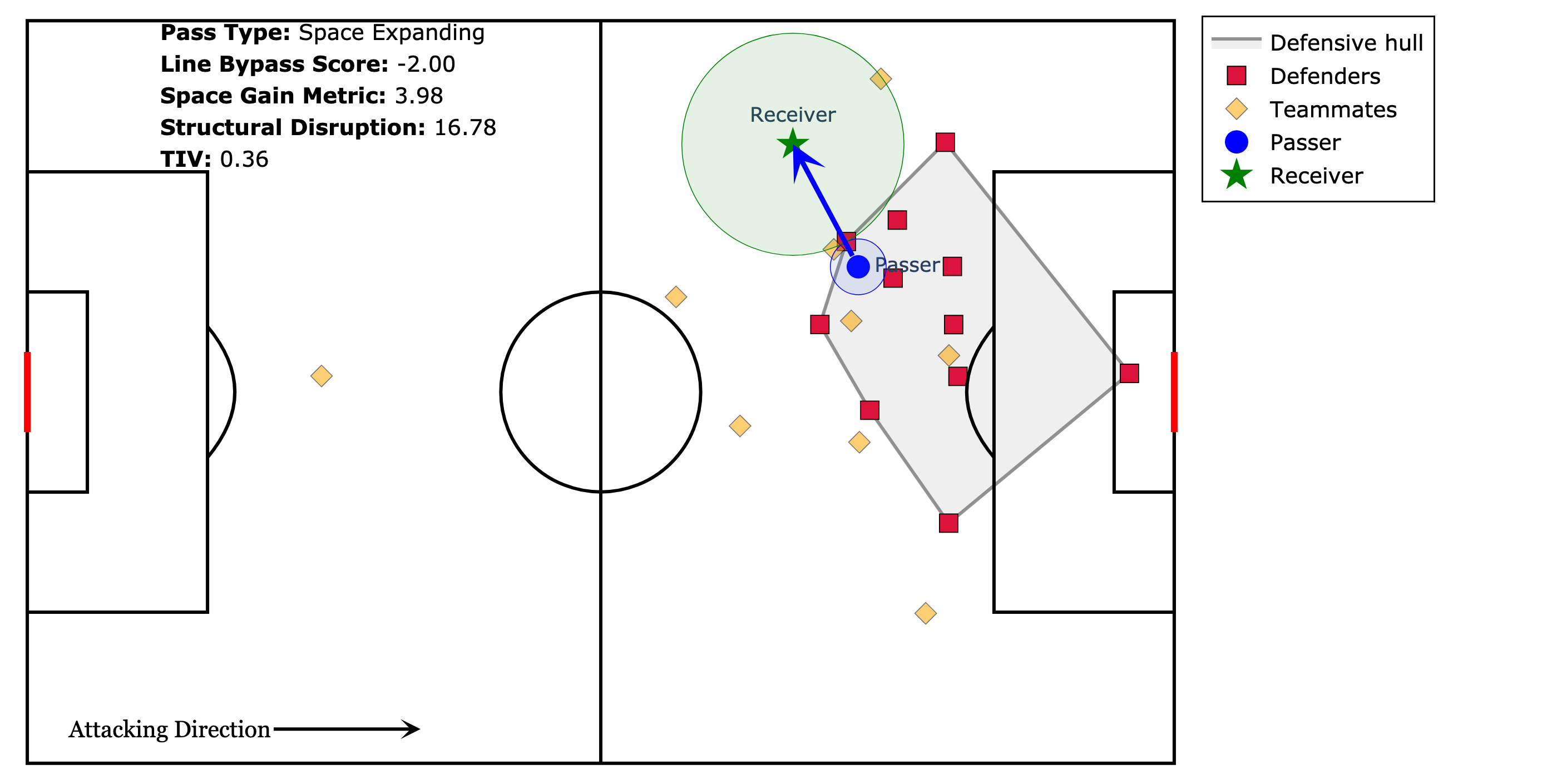}
    \caption{Space-expanding pass}
\end{subfigure}

\caption{Representative examples of the four structural pass archetypes. Each panel shows a pass in tracking context together with the surrounding defensive structure.}
\label{fig:pass_examples}

\end{figure}

Figure~\ref{fig:pass_examples} illustrates representative situations associated with the four archetypes. Circulatory passes exhibit comparatively modest structural measurements, whereas Destabilising passes are characterised by larger differences between sampled defensive configurations. Line-breaking passes display stronger positional bypass, while Space-expanding passes involve greater relative receiving space. The examples illustrate how passes can exhibit different structural characteristics even when they serve broadly similar functions in possession.

\subsection[Structural signature and distribution of pass archetypes]{Structural signature and distribution of pass\\archetypes}

\begin{figure}[t]
    \centering
    \includegraphics[width=\columnwidth]{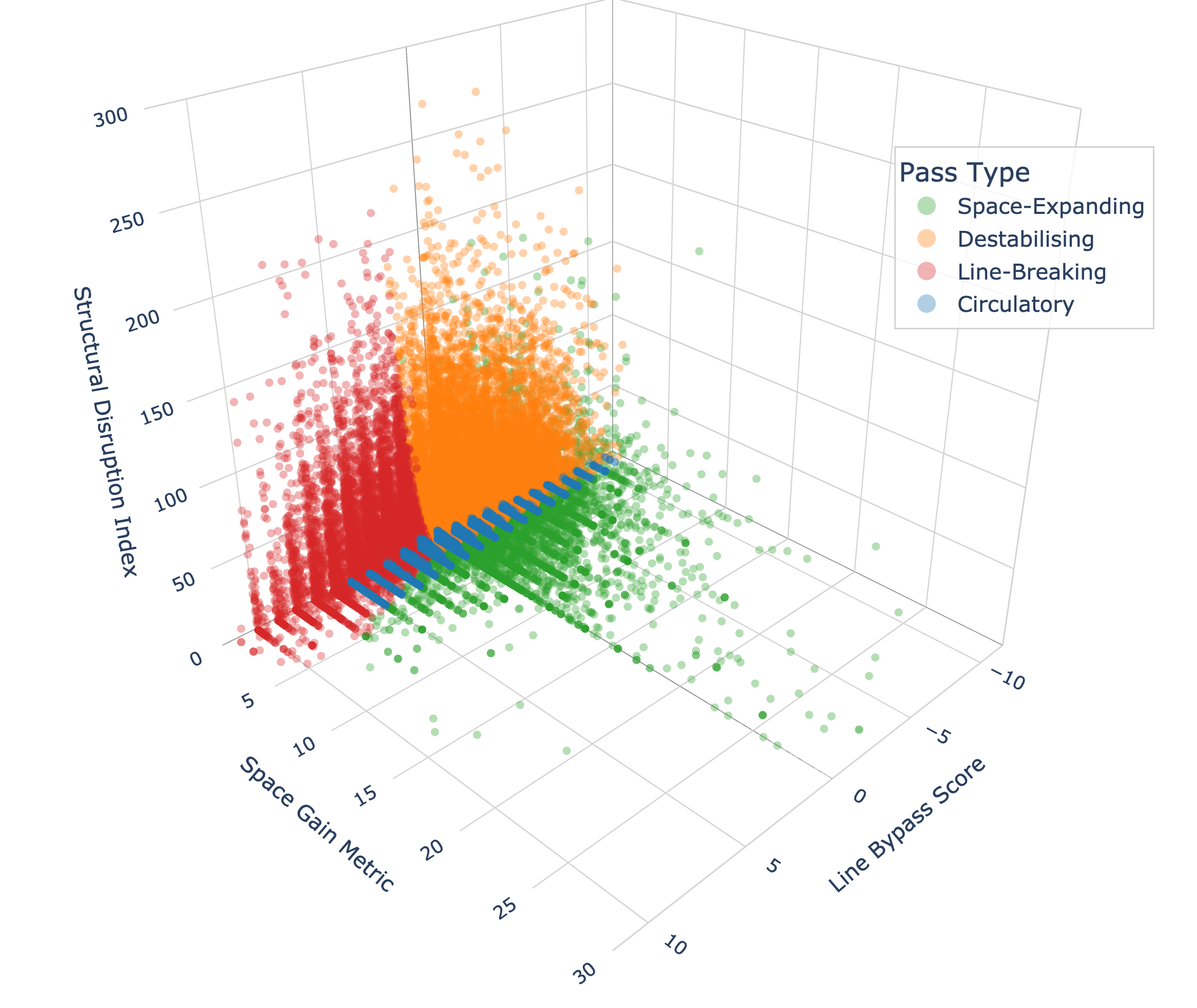}
    \caption{Three-dimensional scatterplot of pass events using the raw Line Bypass Score, Space Gain Metric, and Structural Disruption Index values. Colours indicate assignment to the selected operational pass archetype; the plotted coordinates are not the transformed and standardised features used to fit K-means.}
    \label{fig:structural_map}
\end{figure}

\begin{figure}[t]
    \centering
    \includegraphics[width=\columnwidth]{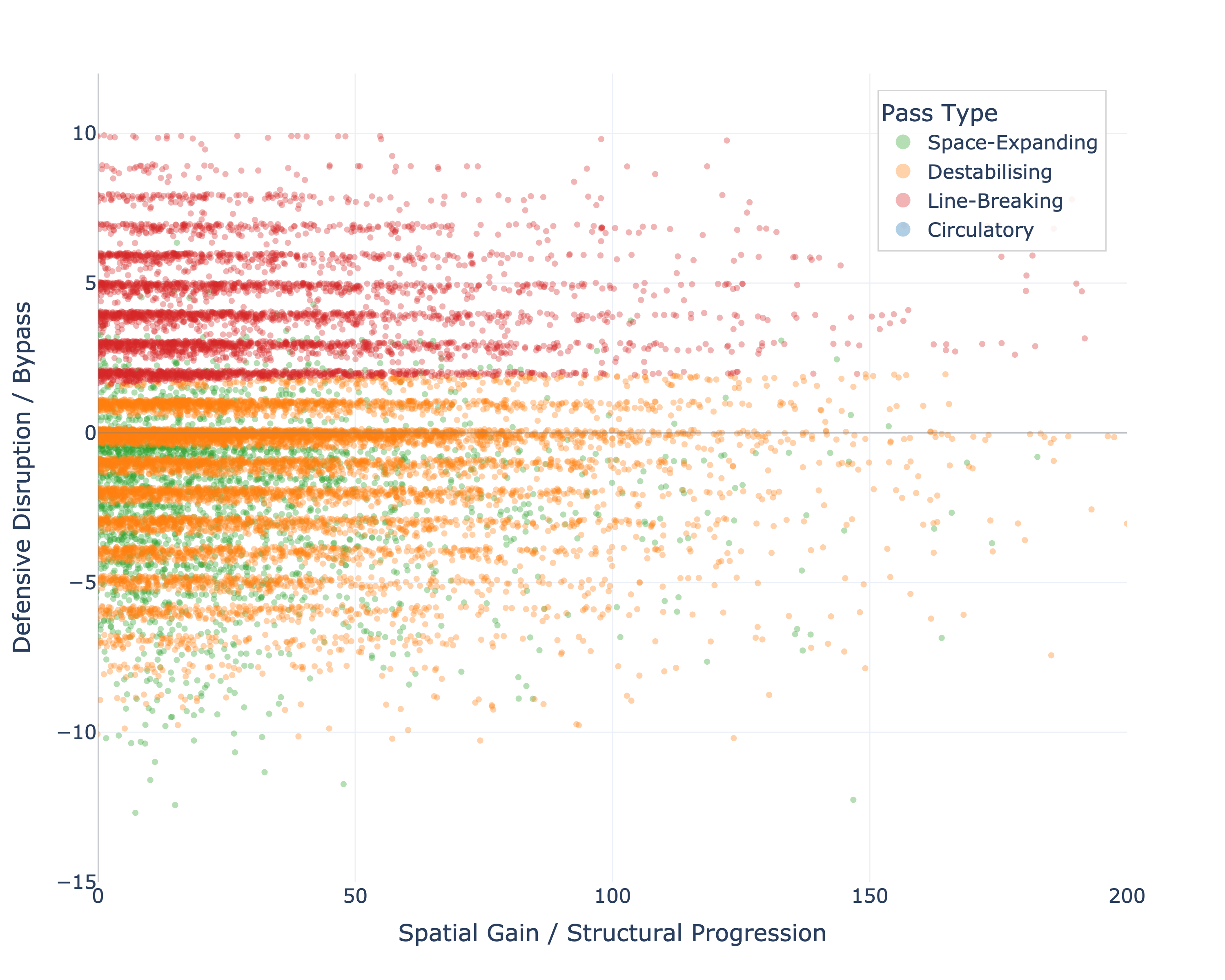}
    \caption{Two-dimensional PCA projection of the three raw structural-metric columns. The plotted component coordinates are descriptive projection scores, not separately measured tactical dimensions; the four operational profiles occupy distinguishable but partially overlapping regions.}
    \label{fig:structural_projection}
\end{figure}

The four archetypes exhibit distinguishable structural signatures, as illustrated by the three-dimensional distribution of raw LBS, SGM, and SDI values in Figure~\ref{fig:structural_map} and their two-dimensional PCA projection in Figure~\ref{fig:structural_projection}. Although the profiles occupy overlapping regions, their characteristic combinations of measurements reveal different relationships between positional bypass, receiving space, and defensive-shape variation.

Table~\ref{tab:cluster_metrics} quantifies these differences. Circulatory passes have comparatively modest mean values across the three descriptors (LBS = 0.10, SGM = 0.97, SDI = 0.26). Destabilising passes exhibit the highest mean SDI (43.06), while Line-breaking passes have the strongest positive mean LBS (4.49), accompanied by a mean SDI of 33.68. Space-expanding passes display the highest mean SGM (4.96), alongside a negative mean LBS (-1.55). This contrast is particularly informative: passes associated with greater relative receiving space do not necessarily bypass more opponents. The negative mean LBS values for Destabilising (-1.36) and Space-expanding passes further illustrate the signed nature of the positional-bypass measure.

\begin{table}[t]
\centering
\caption{Raw arithmetic member means of LBS, SGM, and SDI by operational pass archetype. Values are calculated over assigned pass events and are neither standardised feature means nor K-means centroid coordinates; clustering used transformed and standardised features.}
\label{tab:cluster_metrics}
\small
\resizebox{\columnwidth}{!}{%
\begin{tabular}{lccc}
\toprule
Pass type & \shortstack{Line Bypass\\Score} & \shortstack{Space Gain\\Metric} & \shortstack{Structural Disruption\\Index} \\
\midrule
Circulatory & 0.10 & 0.97 & 0.26 \\
Destabilising & -1.36 & 1.11 & 43.06 \\
Line-breaking & 4.49 & 0.93 & 33.68 \\
Space-expanding & -1.55 & 4.96 & 16.87 \\
\bottomrule
\end{tabular}%
}
\end{table}

The archetype distribution (Table~\ref{tab:pass_type_distribution}) is dominated by Circulatory passes (14,144; 34.43\%) and Destabilising passes (12,836; 31.25\%), which together account for approximately two-thirds of the sample. Line-breaking (7,507; 18.28\%) and Space-expanding passes (6,591; 16.04\%) constitute the remaining observations. Thus, the structural characteristics captured by the taxonomy occur at substantially different frequencies within the tournament.

\begin{table}[ht]
    \centering
    \caption{Distribution of structural pass archetypes.}
    \label{tab:pass_type_distribution}
    \begin{tabular}{lcc}
        \toprule
        Pass type & Count & Percentage \\
        \midrule
        Circulatory & 14144 & 34.43\% \\
        Destabilising & 12836 & 31.25\% \\
        Line-breaking & 7507 & 18.28\% \\
        Space-expanding & 6591 & 16.04\% \\
        \bottomrule
    \end{tabular}
\end{table}

\subsection{Pass archetypes and attacking outcomes}

\begin{figure}[t]
    \centering
    \includegraphics[width=\columnwidth]{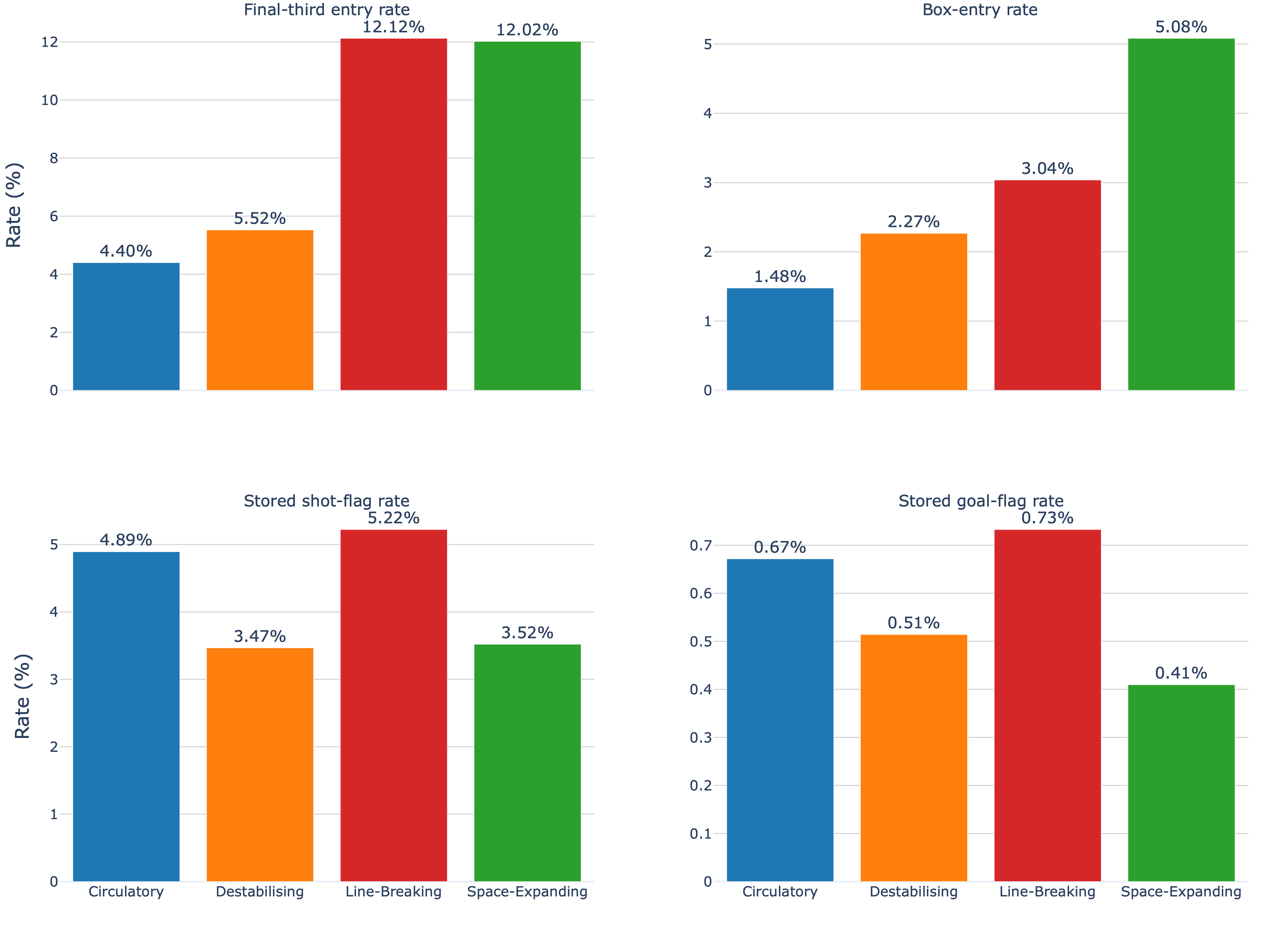}
    \caption{Observed within-archetype rates for direct pass-coordinate final-third and box-entry crossings and for the stored binary shot and goal outcome flags. Territorial crossings describe the pass itself, while the shot and goal columns summarise the corresponding stored indicators.}
    \label{fig:pass_type_effectiveness}
\end{figure}

The four profiles exhibit different relationships with territorial-crossing and recorded attacking indicators (Figure~\ref{fig:pass_type_effectiveness}; Table~\ref{tab:pass_type_outcomes}). Line-breaking and Space-expanding passes have the highest final-third crossing rates, at 12.12\% and 12.02\%, respectively, compared with 5.52\% for Destabilising and 4.40\% for Circulatory passes. However, Space-expanding passes exhibit the highest box-entry rate (5.08\%), followed by Line-breaking (3.04\%), Destabilising (2.27\%), and Circulatory passes (1.48\%).

The recorded attacking indicators reveal a different ordering. Line-breaking passes have the highest stored shot-flag (5.22\%) and goal-flag (0.73\%) rates, followed by Circulatory passes (4.89\% and 0.67\%). Destabilising passes record corresponding rates of 3.47\% and 0.51\%, while Space-expanding passes exhibit 3.52\% and 0.41\%. The contrast between Line-breaking and Space-expanding passes is particularly notable: their final-third crossing rates are almost identical, yet their box-entry and recorded outcome indicators differ. Structural passing characteristics therefore exhibit distinct relationships with the different attacking measures rather than following a single ordering.

\begin{table}[ht]
\centering
\caption{Within-archetype rates for direct pass-coordinate territorial-crossing indicators and stored binary shot and goal flags. Final-third and box entries are defined by coordinate-crossing criteria, while shot and goal rates are calculated from the corresponding stored binary variables.}
\label{tab:pass_type_outcomes}
\small
\resizebox{\columnwidth}{!}{%
\begin{tabular}{lcccc}
\toprule
Pass type & \shortstack{Final-third\\crossing} & \shortstack{Box-entry\\crossing} & \shortstack{Stored shot\\flag} & \shortstack{Stored goal\\flag} \\
\midrule
Circulatory      & 0.043976 & 0.014777 & 0.048925 & 0.006717 \\
Destabilising    & 0.055235 & 0.022671 & 0.034668 & 0.005142 \\
Line-breaking    & 0.121220 & 0.030372 & 0.052218 & 0.007326 \\
Space-expanding  & 0.120164 & 0.050827 & 0.035200 & 0.004096 \\
\bottomrule
\end{tabular}%
}
\end{table}

\subsection{Team-level structural styles}

\begin{figure}[t]
    \centering
    \includegraphics[width=\columnwidth]{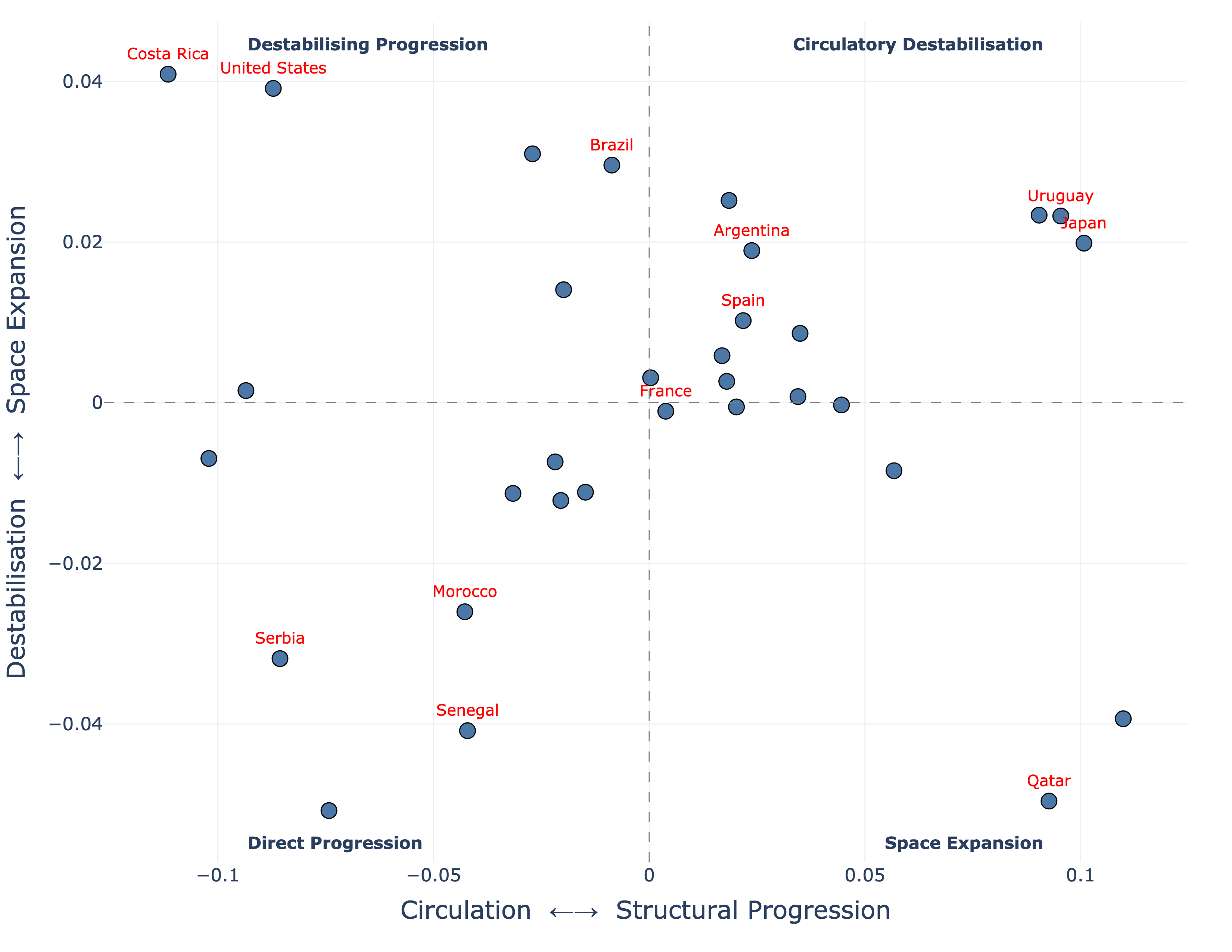}
    \caption{Two-dimensional PCA-based display of team profiles. The axis, quadrant, and colour annotations provide interpretive descriptions of regions within the projection rather than directly measured tactical dimensions or independently verified dominant-archetype classifications.}
    \label{fig:team_style_map}
\end{figure}

Team-level aggregation reveals variation in the composition of structural passing profiles across the tournament. The PCA-based representation in Figure~\ref{fig:team_style_map} places Costa Rica and the United States within the region labelled ``Destabilising Progression'', Japan and Uruguay within ``Circulatory Destabilisation'', Qatar within ``Space Expansion'', and Serbia and Senegal within ``Direct Progression''. These positions provide a comparative view of the teams' observed structural characteristics, with the annotated regions serving as descriptive interpretations of the projection.

\begin{figure}[t]
    \centering
    \includegraphics[width=\columnwidth]{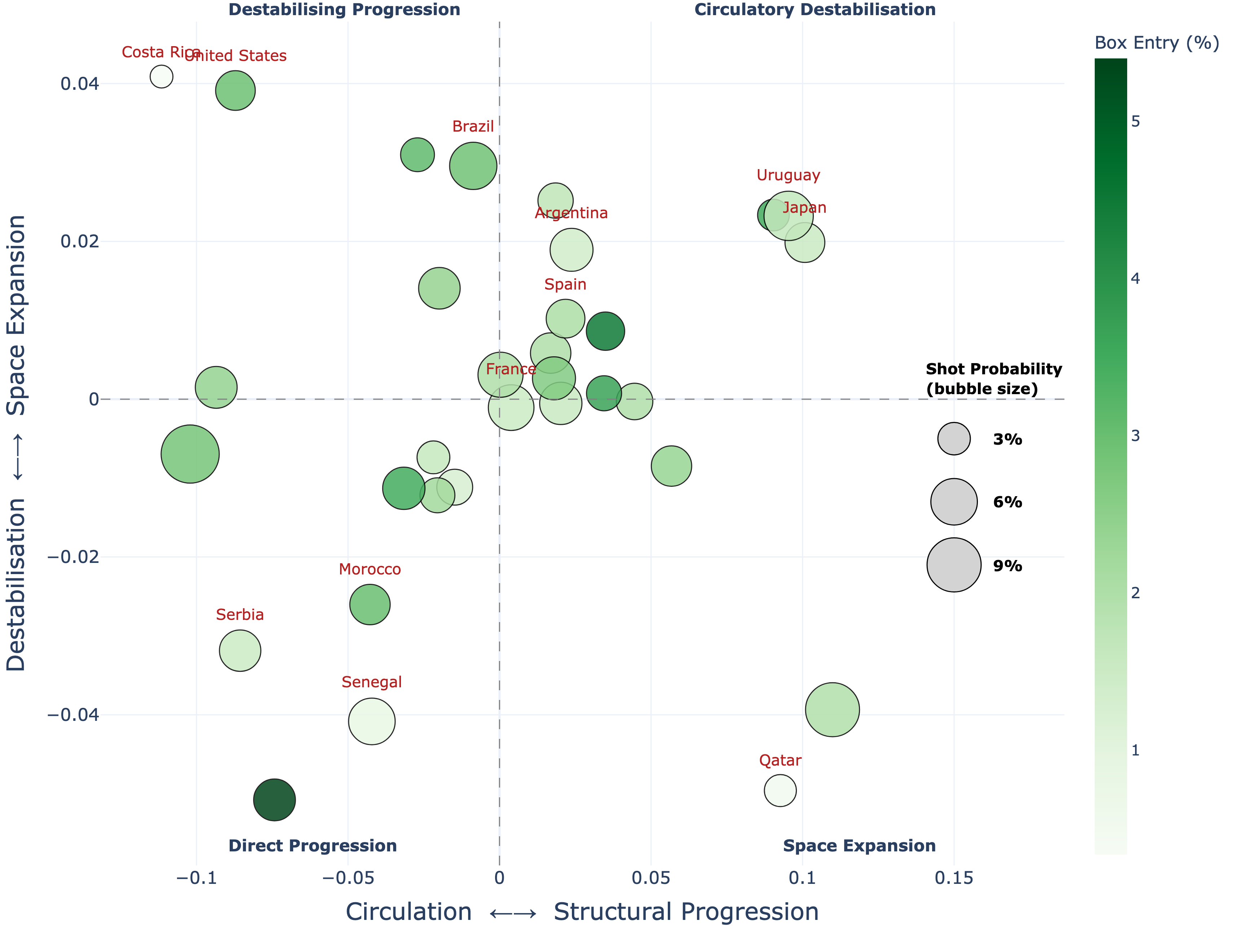}
    \caption{PCA-based team positions with the within-team mean stored shot flag represented by bubble size and the box-entry crossing rate represented by colour. The display combines team-level structural profiles with descriptive attacking indicators.}
    \label{fig:style_vs_productivity}
\end{figure}

Figure~\ref{fig:style_vs_productivity} introduces additional attacking indicators into this representation. Bubble size reflects the within-team mean stored shot flag, while colour indicates the box-entry crossing rate. Together, the two displays allow differences in structural passing profiles to be considered alongside territorial and attacking indicators, providing complementary perspectives on team-level passing behaviour.

\subsection{Tactical Impact Value (\tiv)}

\begin{figure}[t]
    \centering
    \includegraphics[width=\columnwidth]{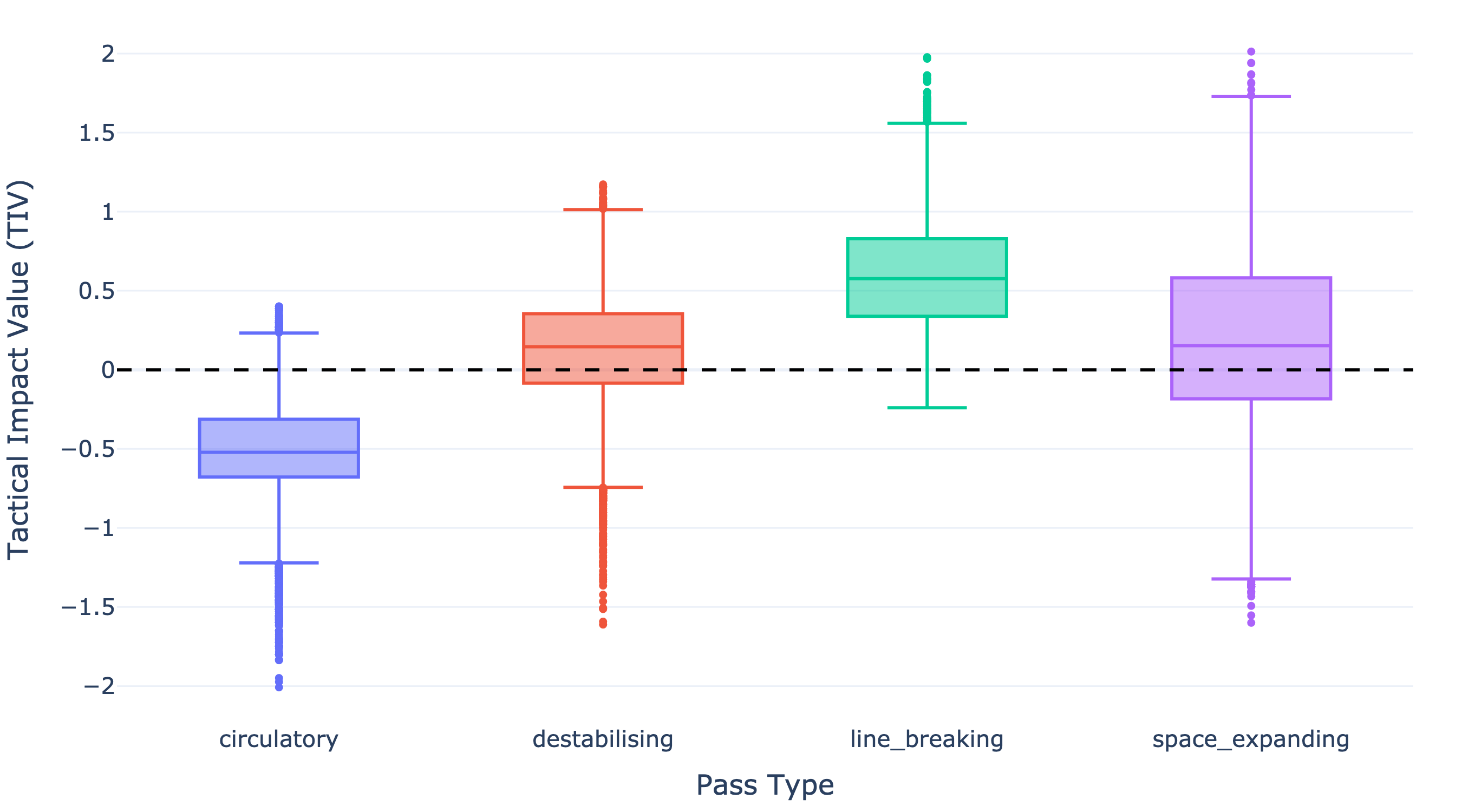}
    \caption{Distribution of Tactical Impact Value (\tiv) across structural pass archetypes.}
    \label{fig:tiv_by_type}
\end{figure}

TIV distributions differ considerably between the four archetypes (Figure~\ref{fig:tiv_by_type}; Table~\ref{tab:tiv_by_type}). Line-breaking passes exhibit the highest mean TIV (0.589), followed by Space-expanding (0.190) and Destabilising passes (0.125), while Circulatory passes have a negative mean (-0.514). Their medians follow the same ordering, at 0.577, 0.153, 0.146, and -0.521, respectively. Space-expanding passes exhibit the greatest variability (standard deviation = 0.539), compared with 0.361 for Line-breaking, 0.350 for Destabilising, and 0.308 for Circulatory passes. These differences reflect the contrasting combinations of structural characteristics represented within each profile.

\begin{table}[ht]
\centering
\caption{Summary statistics of Tactical Impact Value (\tiv) by structural pass type.}
\label{tab:tiv_by_type}
\resizebox{\columnwidth}{!}{%
\begin{tabular}{lccc}
\toprule
Pass type & Mean TIV & Median TIV & Std. Dev. \\
\midrule
Circulatory      & -0.514069 & -0.521473 & 0.308296 \\
Destabilising    &  0.124631 &  0.146360 & 0.349846 \\
Line-breaking    &  0.588821 &  0.576782 & 0.360720 \\
Space-expanding  &  0.189798 &  0.153157 & 0.538823 \\
\bottomrule
\end{tabular}%
}
\end{table}

\begin{figure}[t]
    \centering
    \includegraphics[width=\columnwidth]{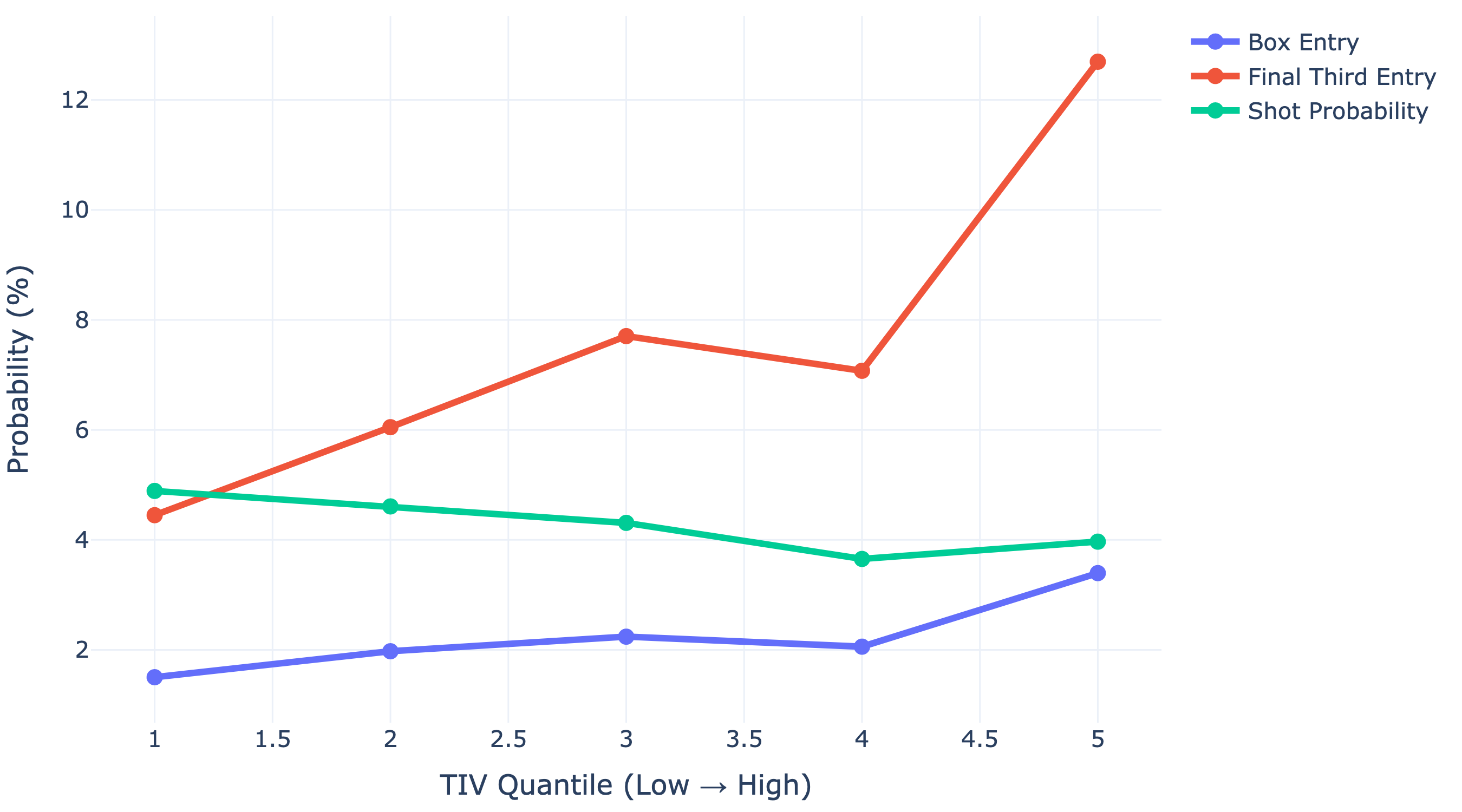}
    \caption{Observed rates across Tactical Impact Value quantiles for direct pass-coordinate final-third and box-entry crossings and the stored binary shot flag. The territorial indicators describe coordinate crossings, while the shot series represents the corresponding stored binary variable.}
    \label{fig:tiv_vs_outcomes}
\end{figure}

Figure~\ref{fig:tiv_vs_outcomes} examines territorial and attacking indicators across five TIV quantiles. Final-third crossing rates increase from approximately 4.5\% in the lowest quantile to more than 12\% in the highest, while box-entry rates rise from approximately 1.5\% to more than 3\%. Neither relationship is strictly monotonic, with both indicators declining between the third and fourth quantiles. The stored shot-flag rate varies more modestly. Thus, higher composite values are associated with greater territorial-crossing frequencies when comparing the extreme quantiles, although the intermediate variation reveals a more complex relationship.

\subsection{Spatial distribution of Tactical Impact Value}

\begin{figure*}[htbp]
    \centering
    \begin{subfigure}[b]{0.48\linewidth}
        \centering
        \includegraphics[width=\linewidth]{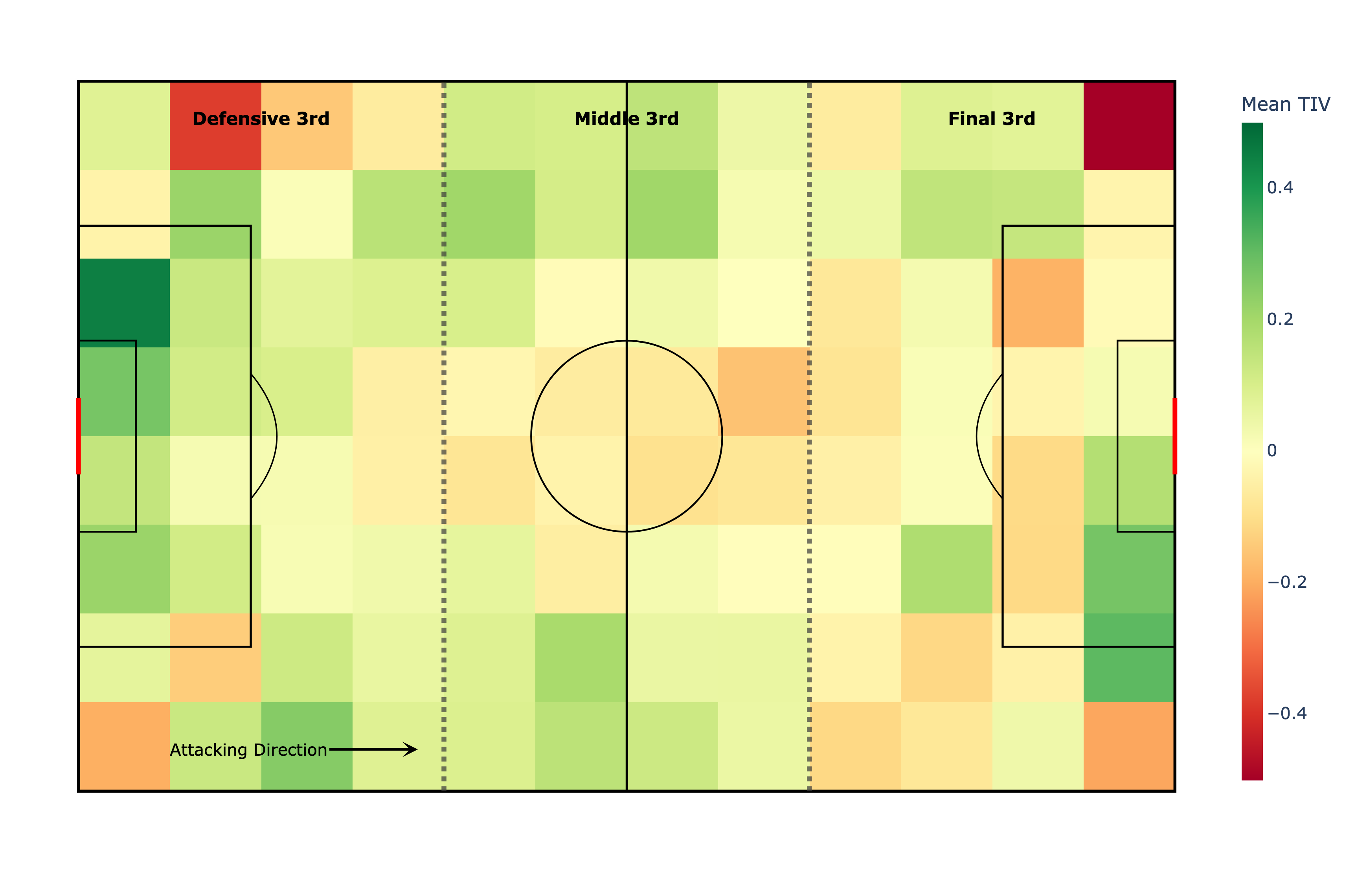}
        \caption{Pass-origin TIV}
        \label{fig:tiv_origin}
    \end{subfigure}
    \hfill
    \begin{subfigure}[b]{0.48\linewidth}
        \centering
        \includegraphics[width=\linewidth]{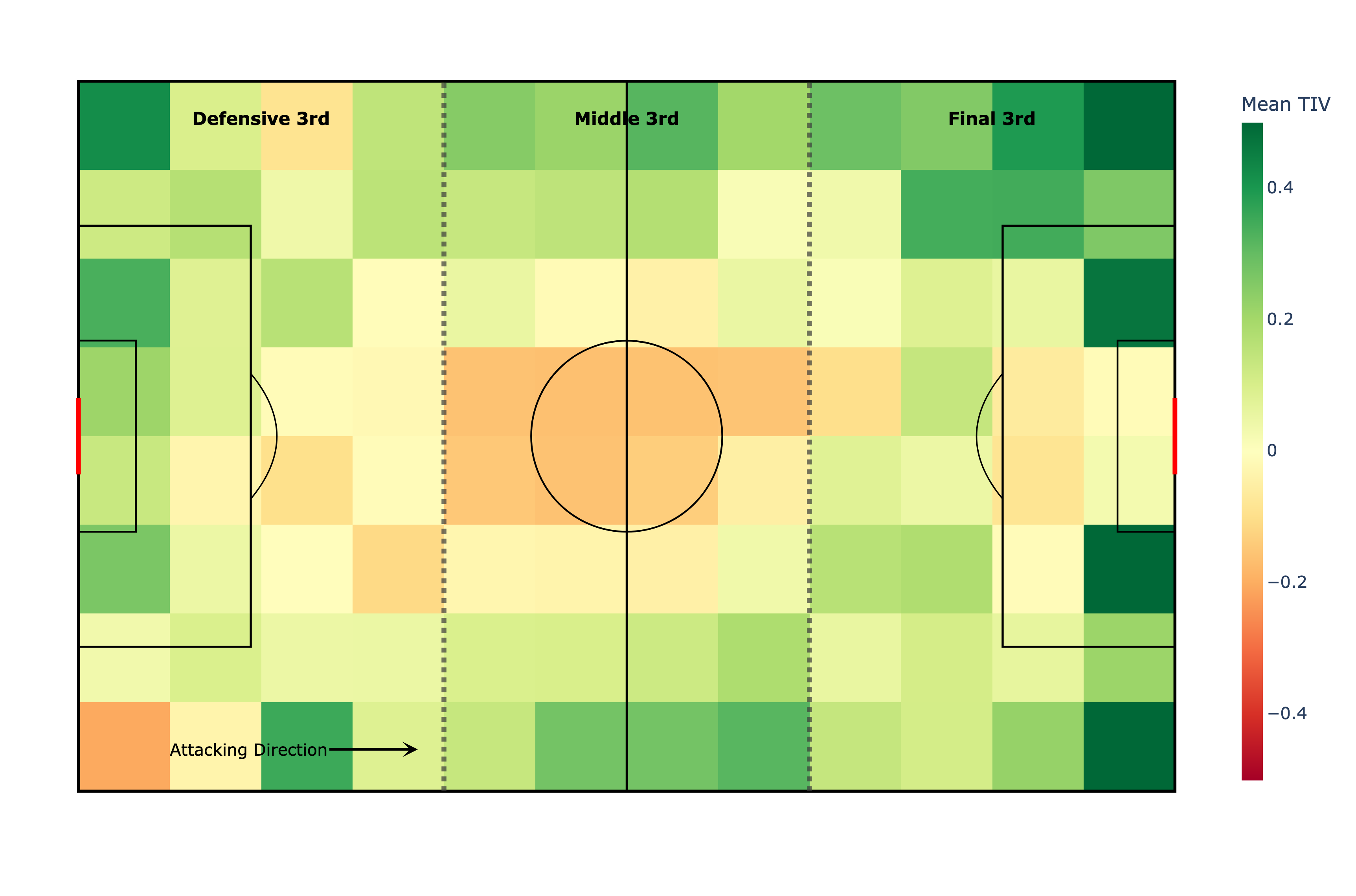}
        \caption{Pass-destination TIV}
        \label{fig:tiv_destination}
    \end{subfigure}
    \caption{Spatial heatmaps of mean Tactical Impact Value by pass location. Panel (a) shows the mean TIV of passes by origin bin, while Panel (b) shows the corresponding mean by destination bin. The maps summarise the coordinate fields used in their construction; their direction-normalisation state is not uniquely established.}
    \label{fig:tiv_spatial}
\end{figure*}

The spatial distribution of TIV is examined through origin- and destination-based heatmaps (Figure~\ref{fig:tiv_spatial}). The former describes mean composite values associated with passes initiated from different coordinate bins, while the latter characterises passes directed towards those locations. Their comparison provides complementary views of where particular structural passing characteristics occur across the tournament. By retaining spatial information alongside the composite, these representations support the identification of areas warranting more detailed examination of the underlying descriptors.

\subsection{Comparative team TIV maps}

\begin{figure*}[ht]
    \centering
    \includegraphics[width=\linewidth]{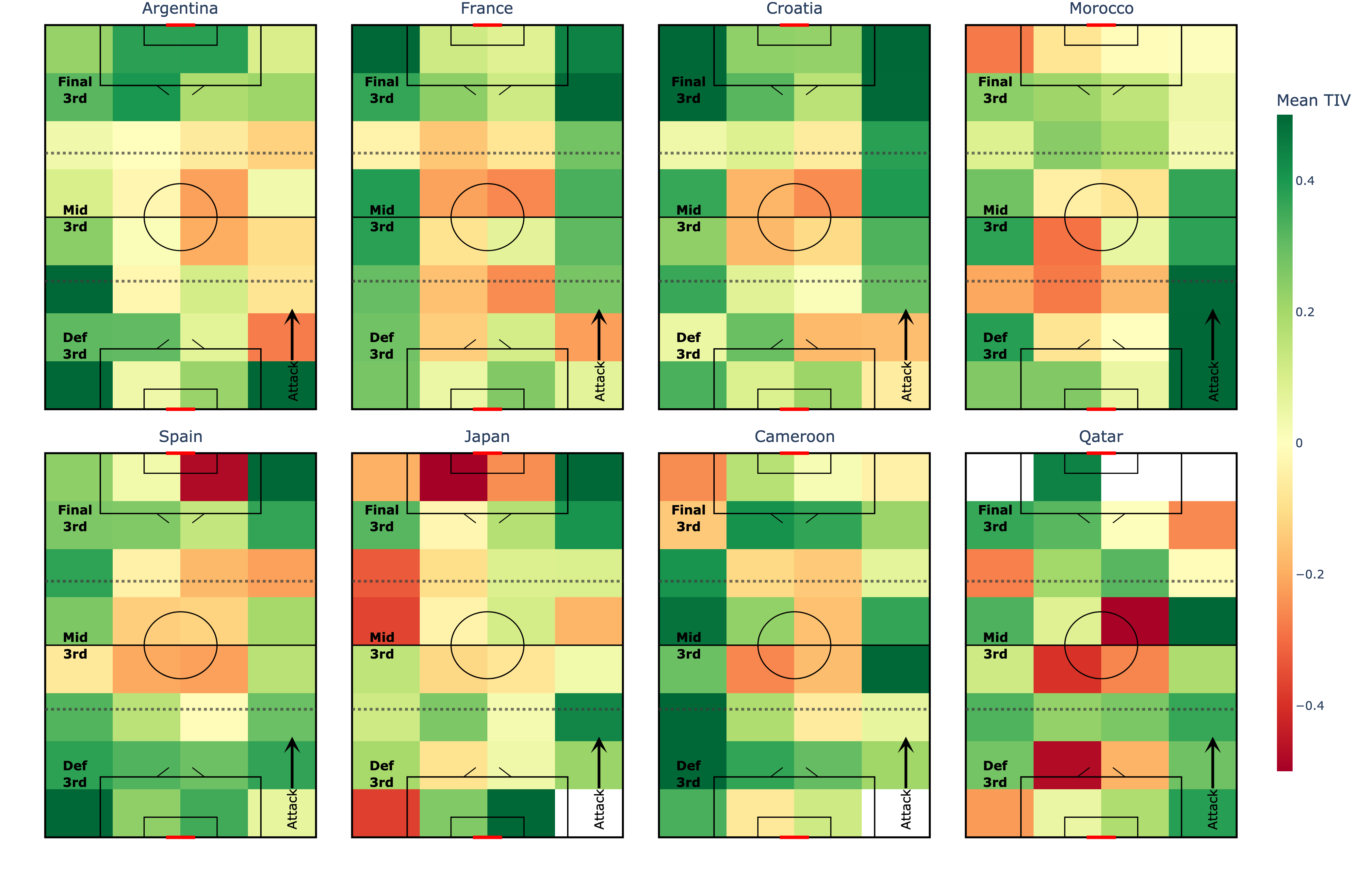}
    \caption{Team-level maps of mean destination Tactical Impact Value for eight selected World Cup teams. Colour represents the mean structural-composite value within each displayed coordinate bin. The exact direction-normalisation state of the mapped coordinates is not uniquely established.}
    \label{fig:team_tiv_maps}
\end{figure*}

Figure~\ref{fig:team_tiv_maps} extends the spatial analysis to eight selected national teams, including the four semi-finalists. The destination-based heatmaps reveal differences in the spatial distribution of mean TIV across the displayed coordinate bins. Whereas the team-level PCA representation summarises structural passing profiles, these maps retain information about where passes with different composite values are directed. The two approaches consequently offer complementary descriptions of team passing behaviour, combining its structural composition with its spatial distribution.

\subsection{Player-level structural passing profiles}

\begin{figure*}[ht]
    \centering
    \includegraphics[width=\linewidth]{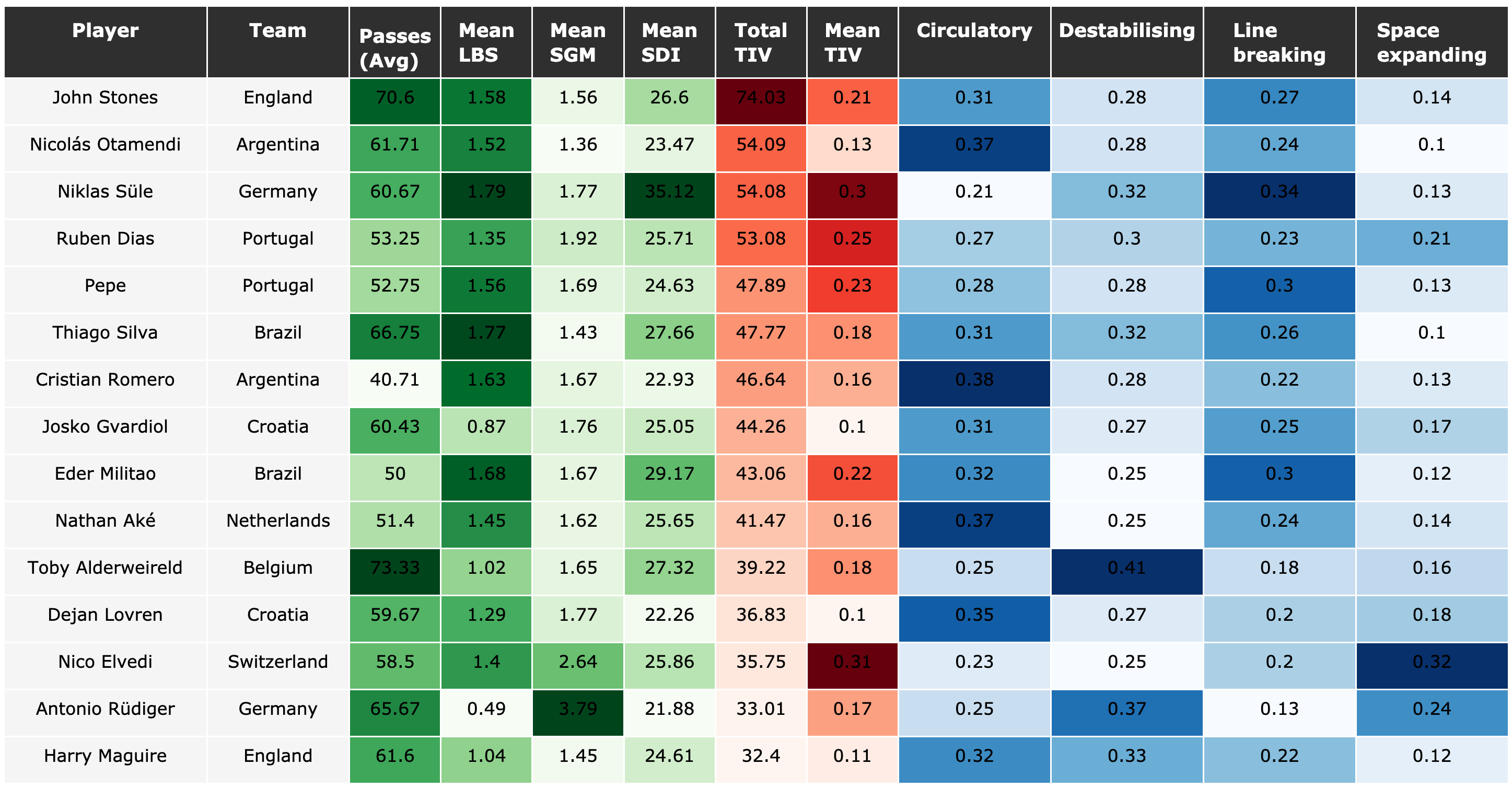}
    \caption{Player-level structural passing profiles for the 15 displayed players ranked by cumulative Tactical Impact Value (TIV), after applying a minimum threshold of 40 mean analysed passes per match. ``Passes (Avg)'' is the mean pass count per match. The table reports mean structural metrics, cumulative and mean TIV, and the proportions of passes in each operational archetype. Cumulative TIV reflects both pass-level values and the number of analysed passes, whereas mean TIV is the average per analysed pass.}
    \label{fig:player_structural_profiles}
\end{figure*}

Figure~\ref{fig:player_structural_profiles} presents the 15 players with the highest cumulative TIV among those averaging at least 40 analysed passes per match. Several central defenders feature prominently, including John Stones, Nicol\'as Otamendi, Niklas S\"ule, R\'uben Dias, and Thiago Silva. Their inclusion illustrates the relevance of structural passing characteristics across different areas and phases of possession.

The component metrics reveal important differences within this group. S\"ule and \'Eder Milit\~ao exhibit comparatively high mean LBS values, whereas Nico Elvedi and Antonio R\"udiger display relatively high mean SGM values. The archetype proportions provide further differentiation: Circulatory passes are common across the displayed players, S\"ule and Milit\~ao have comparatively larger Line-breaking shares, and Toby Alderweireld exhibits a larger Destabilising share. These patterns demonstrate that players with substantial cumulative TIV can possess contrasting structural passing profiles. Cumulative TIV reflects both pass volume and composite values, while mean TIV describes the average per analysed pass.

Passer--receiver comparisons provide an additional perspective on individual passing behaviour. For each combination, the pair's mean TIV is compared with the passer's overall mean:

\[
\Delta TIV_{p \rightarrow r} =
\overline{TIV}_{p \rightarrow r} - \overline{TIV}_{p}.
\]

\begin{figure*}[ht]
    \centering
    \includegraphics[width=\linewidth]{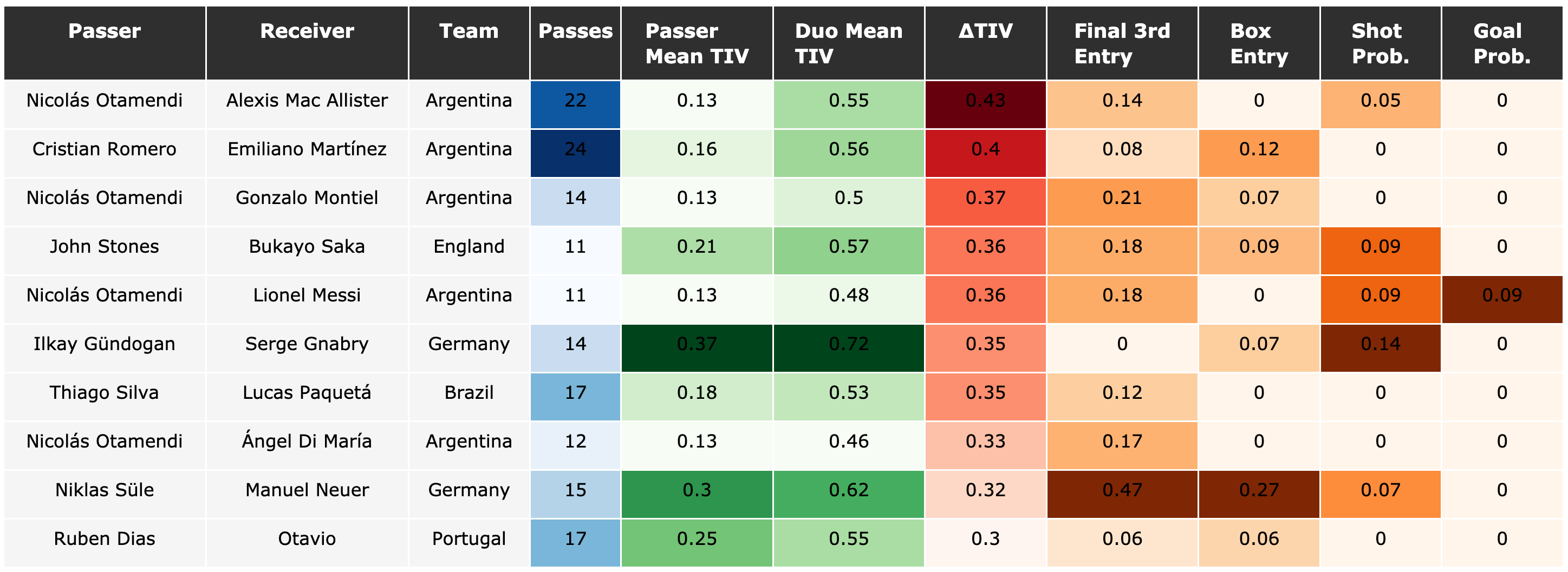}
    \caption{The 10 displayed passer--receiver pairs with the largest $\Delta TIV$ among pairs from the 10 passers with the highest cumulative TIV, with a minimum of 10 passes per pair. $\Delta TIV$ is the difference between pair mean TIV and the passer's mean TIV across all analysed passes. Shot and goal columns report within-pair proportions of stored binary flags, while territorial entries represent the corresponding stored-coordinate crossing indicators.}
    \label{fig:passing_duos_tiv}
\end{figure*}

Positive values indicate that passes to a particular receiver exhibit higher average structural-composite values than the passer's overall reference. Figure~\ref{fig:passing_duos_tiv} displays the ten combinations with the largest positive differences among eligible pairs involving the ten highest-ranked passers by cumulative TIV, requiring at least ten analysed passes per pair.

Several combinations involve Nicol\'as Otamendi, including exchanges with Alexis Mac Allister, Gonzalo Montiel, and Lionel Messi. Other examples include John Stones--Bukayo Saka, Ilkay G\"undogan--Serge Gnabry, and Thiago Silva--Lucas Paquet\'a. The Niklas S\"ule--Manuel Neuer combination is distinctive in its territorial indicators, with approximately 47\% final-third crossings, 27\% box entries, and a 7\% stored shot-flag rate. Together, these comparisons illustrate how the structural characteristics of a player's passing can vary across receiving relationships, providing a more detailed account than player-level averages alone.

\section{Discussion}
\label{sec:discussion}

The findings demonstrate how passing behaviour can be examined through its relationship with opposition positioning and defensive organisation. Across 41,078 open-play passing events from the 2022 FIFA World Cup, the proposed descriptors reveal structural differences that are not fully represented by pass locations or attacking outcomes alone. Their combination into four interpretable archetypes provides a common framework for examining these differences across individual actions, teams, players, and recurring passing relationships.

A central finding is that positional bypass, relative receiving space, and defensive-shape variation capture distinct characteristics of passing. Line-breaking passes exhibit the strongest positive positional-bypass signature, whereas Space-expanding passes are distinguished by larger nearest-opponent distance ratios. Destabilising passes display greater differences between sampled defensive configurations, while Circulatory passes have comparatively modest structural measurements. These distinctions are meaningful because passes can bypass several opponents without substantially increasing receiving space, or reach relatively unmarked receivers without bypassing many defenders. Defensive-shape variation introduces a further dimension that cannot be inferred directly from either positional relationship. The resulting taxonomy therefore describes recurring combinations of spatial characteristics rather than alternative expressions of a single underlying measure.

The relationships with territorial and attacking indicators reinforce this distinction. Line-breaking and Space-expanding passes exhibit similarly high final-third crossing rates, yet Space-expanding passes have the highest penalty-area entry rate. Conversely, Line-breaking passes record the highest stored shot- and goal-flag rates, while Circulatory passes also exhibit a comparatively high shot-flag rate. These differences suggest that the structural characteristics associated with territorial progression are not necessarily those most frequently associated with the recorded attacking indicators. Their interpretation remains descriptive, particularly given the unresolved derivation of the stored outcome variables.

Tactical Impact Value (TIV) offers a complementary summary of these characteristics. Its distributions distinguish the four archetypes, with Line-breaking passes exhibiting the highest mean and Circulatory passes the lowest. However, the considerable variation within profiles, particularly Space-expanding passes, illustrates why the composite should be interpreted alongside its constituent descriptors. The higher territorial-crossing rates observed in the highest TIV quantile, together with the non-monotonic variation across intermediate quantiles, further indicate that structural measurements and territorial progression have a relationship that is more complex than a simple ranking. TIV is therefore most informative as a compact representation of measured passing structure rather than a substitute for outcome-oriented action valuation.

The analyses across different scales demonstrate the framework's practical flexibility. Team-level archetype proportions provide a basis for comparing observed passing characteristics, while spatial TIV maps retain information about where these characteristics occur. Player-level results reveal contrasting structural profiles among individuals with high cumulative TIV, including several central defenders. Because cumulative TIV depends on both passing volume and individual composite values, comparisons benefit from considering mean TIV, component metrics, and archetype distributions together. Passer--receiver analysis adds a relational perspective by identifying combinations whose mean structural values differ from the passer's overall reference. Collectively, these analyses offer complementary ways to investigate passing behaviour without reducing it to a single player or team ranking.

The framework also has a direct application to football analysis. Positional bypass can help identify passes that reach receivers beyond greater numbers of opponents; relative-space measurements can highlight receiving situations with different levels of immediate defensive proximity; and defensive-shape comparisons can direct attention towards actions accompanied by substantial changes in opposition organisation. These measurements provide starting points for tactical investigation, particularly when combined with event context, tracking visualisations, and video review.

This contribution complements established research on passing decisions and receiver selection \cite{rahimian2023passreceiver,rahimian2025pass}, as well as studies of penetrative passing and defensive organisation \cite{rahimian2022penetrative}. Whereas these approaches investigate passing opportunities, execution, or particular forms of penetration, the present framework offers a multidimensional description of observed passing structure. Its principal contribution lies in connecting interpretable tracking-derived measurements with recurring patterns of football behaviour, while retaining sufficient detail to investigate the spatial characteristics underlying those patterns.

\section{Limitations and Future Work}
\label{sec:limitations}

Several considerations define the scope of the findings. First, the sample comprises identified open-play passing events rather than independently verified completed passes. The analysis consequently describes observed passing situations without accounting for unsuccessful execution or the decision process through which a pass was selected. Future work should distinguish completion outcomes and investigate how structural characteristics interact with passing feasibility and risk.

Each descriptor also involves specific measurement assumptions. LBS compares goal-side opponent counts at pass initiation without reconstructing defensive lines or tracking individual defender movements. SGM represents relative nearest-opponent distance rather than comprehensive spatial control, and its denominator floor of 1.5 requires sensitivity analysis. SDI compares defensive-shape summaries at pass initiation and approximately one second later, independently of reception time or possession continuity. The resulting differences cannot be attributed causally to the pass. Furthermore, its five components retain their original numerical scales, potentially contributing unequally to the unnormalised aggregate. Alternative temporal definitions, component scaling, and spatial-control measures warrant investigation.

TIV introduces fixed weights of 0.35, 0.25, and 0.40 for LBS, SGM, and SDI, respectively. These weights were not empirically optimised or externally validated. Similarly, the four-profile taxonomy reflects the selected K-means configuration and feature transformations rather than an established natural partition of passing behaviour. Sensitivity analyses involving alternative weights, transformations, clustering methods, and numbers of profiles would help establish the robustness and interpretability of both representations.

The additional attacking indicators require particular caution. Although territorial entries follow specified coordinate-crossing rules, the processing record does not uniquely establish whether their coordinates were direction-normalised. Their interpretation as forward territorial progression therefore remains conditional. The generation lineage of the stored shot and goal flags is also unresolved, preventing definitive interpretation of their event windows or associated outcomes. These variables are consequently used descriptively. Reconstructing their derivation, confirming the upstream tracking-smoothing procedure, and verifying correspondence between the processing pipeline and reported outputs are priorities for reproducibility.

Finally, the analysis concerns a single international tournament. Its structural distributions and player-level patterns may differ across domestic competitions, seasons, tactical systems, and match contexts. External validation should examine the stability of the descriptors and archetypes across more diverse datasets. Integrating structural measurements with established action-value models, contextual passing information, and systematic video analysis would further clarify their complementary analytical contribution and practical relevance.

\section{Conclusion}
\label{sec:conclusion}

This study developed an interpretable framework for examining football passing behaviour through its relationship with opposition positioning and defensive organisation. Three complementary descriptors characterise positional bypass, relative receiving space, and differences between sampled defensive configurations. Their integration into Tactical Impact Value (TIV) and an unsupervised passing taxonomy provides a multidimensional representation of structural passing characteristics.

Applied to 41,078 open-play passes from the 2022 FIFA World Cup, the framework identified four distinct operational profiles: Circulatory, Destabilising, Line-breaking, and Space-expanding. These profiles exhibited different structural signatures and relationships with territorial-crossing and recorded attacking indicators, illustrating how passes can be distinguished beyond their event locations and subsequent outcomes.

Team, spatial, player, and passer--receiver analyses demonstrated the framework's applicability across different analytical scales. By retaining interpretable structural descriptors alongside composite and categorical representations, the approach complements established action-valuation methods and provides a foundation for more detailed investigations of passing behaviour. Further validation across competitions and tactical contexts will help establish its broader analytical and practical relevance.

\FloatBarrier

\section*{Statements and Declarations}

\subsection*{Ethical considerations}

This study used an existing, publicly accessible football tracking and event dataset released by PFF FC. It involved no participant recruitment, interventions, or collection of new participant data. Ethical approval was not required for this secondary-data analysis.

\subsection*{Consent to participate}

Not applicable.

\subsection*{Consent for publication}

Not applicable.

\subsection*{Declaration of conflicting interest}

The author(s) declared no potential conflicts of interest with respect to the research, authorship, and/or publication of this article.

\subsection*{Funding}

The authors received no financial support for the research, authorship, and/or publication of this article.

\subsection*{Data availability}

The tracking and event data analysed in this study were released by PFF FC as part of its 2022 FIFA World Cup dataset. The original data are accessible through the official PFF FC release:

\url{https://www.blog.fc.pff.com/blog/pff-fc-release-2022-world-cup-data}

The authors do not redistribute the original dataset.

\subsection*{Code availability}

The analysis code used in this study is not publicly available and will not be shared.

\subsection*{Author contributions}

Oktay Karaku\c{s}: Conceptualization; Methodology; Software; Formal analysis; Investigation; Data curation; Visualization; Writing -- original draft; Writing -- review \& editing; Project administration.

Hasan Arkada\c{s}: Conceptualization; Methodology; Writing -- original draft; Writing -- review \& editing.

Can Ey\"upo\u{g}lu: Conceptualization; Methodology; Writing -- review \& editing.

\subsection*{Generative AI use}

During manuscript revision, the authors used OpenAI ChatGPT and GitHub Copilot for language refinement, consistency checking, methodological-documentation reconciliation, and journal formatting. The authors reviewed all resulting changes and retain full responsibility for the work's accuracy and integrity. These tools were not used to generate or modify the empirical dataset, alter reported results, or determine scientific conclusions.

\bibliographystyle{SageV}
\bibliography{cas-refs}

\end{document}